\documentclass{article}

\usepackage{arxiv}

\usepackage[utf8]{inputenc} 
\usepackage[T1]{fontenc}    
\usepackage{hyperref}       
\usepackage{url}            
\usepackage{booktabs}       
\usepackage{amsfonts}       
\usepackage{nicefrac}       
\usepackage{microtype}      
\usepackage{lipsum}
\usepackage{graphicx}
\usepackage{subcaption}
\usepackage{mathtools}
\usepackage[export]{adjustbox}
\usepackage{algorithm}
\usepackage{algpseudocode}
\usepackage{xcolor}
\graphicspath{ {./images/} }

\title{Mixture of Diffusers for scene composition and high resolution image generation}

\author{
 Álvaro Barbero Jiménez \\
  Instituto de Ingeniería del Conocimiento and Universidad Autónoma de Madrid\\
  Madrid, Spain \\
  \texttt{alvaro.barbero@iic.uam.es} \\
  \url{https://github.com/albarji/mixture-of-diffusers}
}

\begin{document}
\maketitle

\begin{figure}[h]
    \begin{subfigure}[b]{1.0\textwidth}
        \centering
        \includegraphics[width=\textwidth]{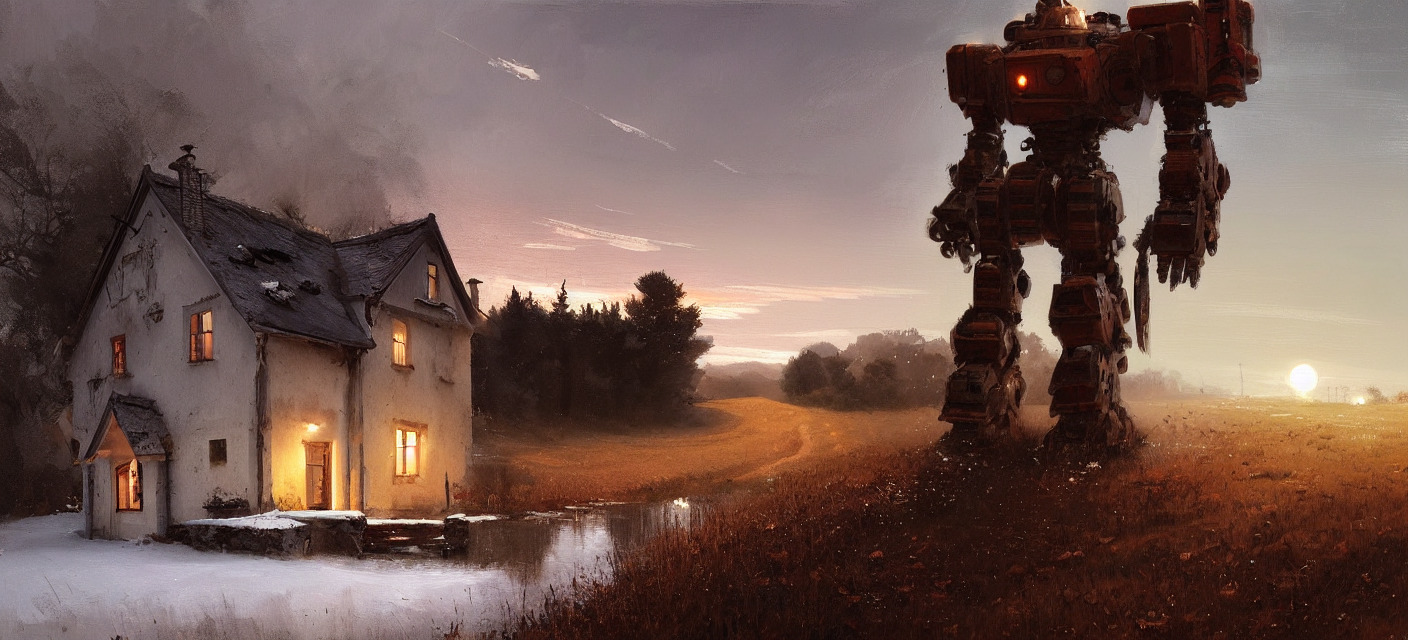}
        \caption{Mixture of Diffusers}
        \label{fig:modTitle_MoD}
    \end{subfigure}
    \begin{subfigure}[b]{1.0\textwidth}
        \centering
        \includegraphics[width=0.33\textwidth]{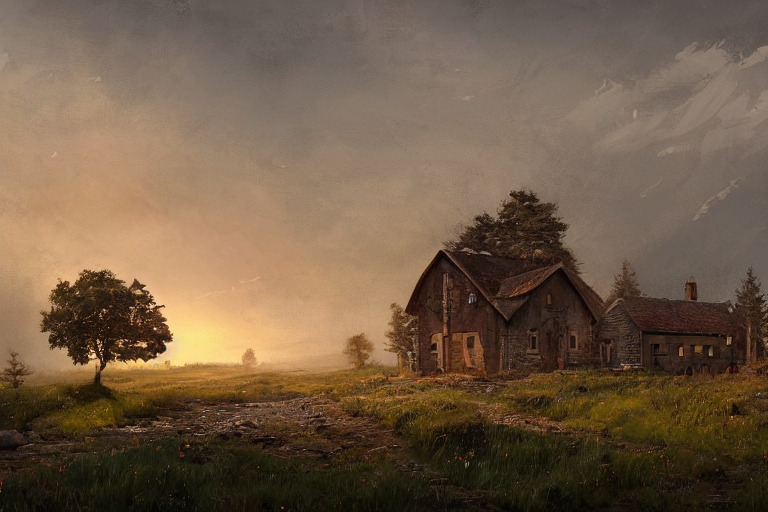}
        \includegraphics[width=0.33\textwidth]{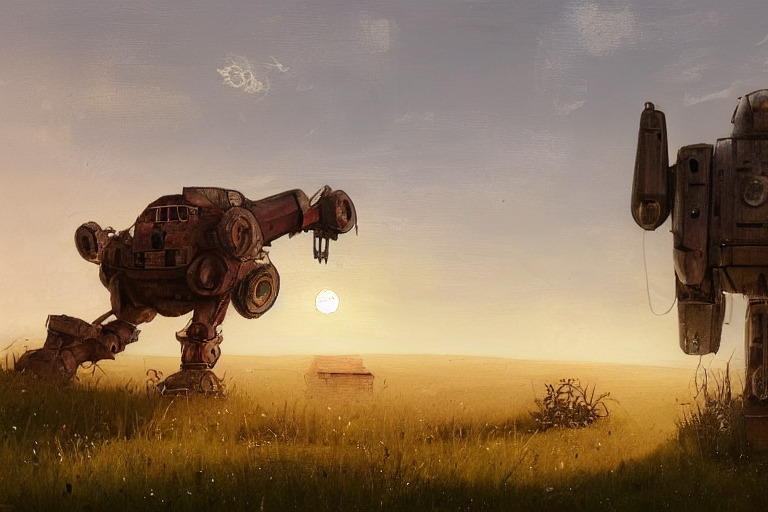}
        \includegraphics[width=0.33\textwidth]{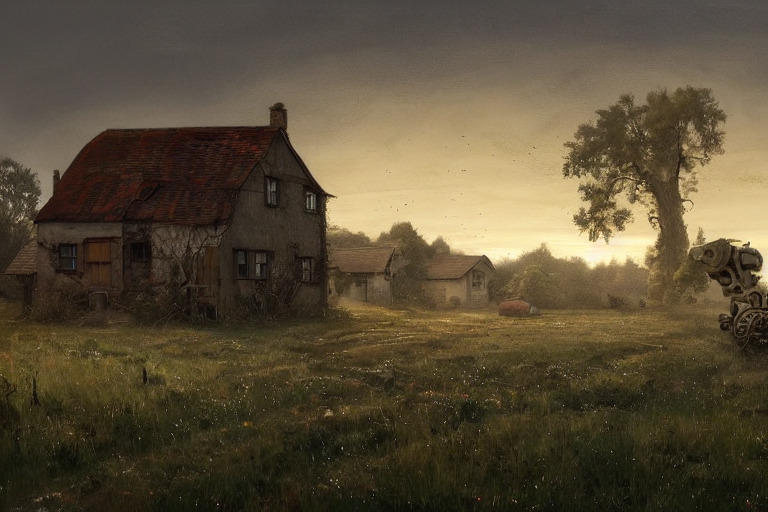}
        \caption{Stable Diffusion}
        \label{fig:modTitle_SD}
    \end{subfigure}
    \caption{An example comparing the image composition capabilities of Mixture of Diffusers and Stable Diffusion.}
    \label{fig:modTitle}
\end{figure}

\begin{abstract}
Diffusion methods have been proven to be very effective to generate images while conditioning on a text prompt. However, and although the quality of the generated images is unprecedented, these methods seem to struggle when trying to generate specific image compositions. In this paper we present Mixture of Diffusers, an algorithm that builds over existing diffusion models to provide a more detailed control over composition. By harmonizing several diffusion processes acting on different regions of a canvas, it allows generating larger images, where the location of each object and style is controlled by a separate diffusion process.
\end{abstract}


\section{Introduction}

Diffusion models \cite{sohl2015diffusionmodels} have proven to be a very effective way to generate synthetic images, be it in an unconditioned way by following a particular training distribution \cite{ho2020denoising, nichol2021improved}, conditioned on a particular class through an auxiliary classifier \cite{dhariwal2021diffusion}, or using a text prompt as a guidance in the generation process \cite{nichol2021glide, ho2022classifierfreeguidance, ramesh2022hierarchical, saharia2022photorealistic, rombach2022high}. They have also been shown to be simpler to train and to generate higher quality images than previous methods based on Generative Adversarial Networks \cite{goodfellow2020generative}.

Nevertheless, text-conditioned diffusion models still have room for improvement in translating the user intent into faithful image representations. Generally, while the semantics of the elements described in a scene seem to be represented to a reasonable degree of accuracy, syntactic relationships between two or more elements are usually represented somewhat vaguely. For instance, Figure \ref{fig:modTitle_SD} shows 3 out of 20 generations using Stable Diffusion 1.4 \cite{rombach2022high} that best represent the \textbf{Unique} prompt in Figure \ref{fig:compositionPrompts}. While the prompt provides instructions for the picture composition, locating specific objects in particular positions, these seem to be mostly ignored in the generated images.

\begin{figure}[h]
    \noindent\fbox{%
        \parbox{\textwidth}{%
            \textbf{Unique}: A charming house in the countryside \textbf{on the left}, \textbf{in the center} a dirt road in the countryside crossing pastures, \textbf{on the right} an old and rusty giant robot lying on a dirt road, by jakub rozalski, sunset lighting on the left and center, dark sunset lighting on the right elegant, highly detailed, smooth, sharp focus, artstation, stunning masterpiece
        }%
    }
    \noindent\fbox{%
        \parbox{\textwidth}{%
            \textbf{Left}: A charming house in the countryside, by jakub rozalski, sunset lighting, elegant, highly detailed, smooth, sharp focus, artstation, stunning masterpiece.
        }%
    }
    \noindent\fbox{%
        \parbox{\textwidth}{%
            \textbf{Center}: A dirt road in the countryside crossing pastures, by jakub rozalski, sunset lighting, elegant, highly detailed, smooth, sharp focus, artstation, stunning masterpiece.
        }%
    }
    \noindent\fbox{%
        \parbox{\textwidth}{%
            \textbf{Right}: An old and rusty giant robot lying on a dirt road, by jakub rozalski, dark sunset lighting, elegant, highly detailed, smooth, sharp focus, artstation, stunning masterpiece.
        }%
    }
    \caption{Prompts used in image composition experiments.}
    \label{fig:compositionPrompts}
\end{figure}

Another known limitation of diffusion models is their scalability to high-resolution images. At its core, a diffusion model requires estimating some form of noise distribution conditioned on a partially denoised image, a procedure generally carried out by a deep neural network, such as U-Net \cite{ronneberger2015u} improved with attention layers \cite{vaswani2017attention, ho2020denoising}. While effective, such network imposes a computational cost in the order of $O(N^2 \log N^2)$ for an $N \times N$ pixels image. But more importantly, the same requirements scaling applies for GPU RAM consumption, thus making the generation of high-resolution images directly out of the diffusion process impractical. In order to overcome this problem several techniques have been proposed, such as upscaling with another diffusion model \cite{ramesh2022hierarchical} or running the diffusion process in a latent space from which pixels can be recovered through a decoder \cite{rombach2022high, stableDiffusion}. Image super-resolution models based on GANs can be applied as well \cite{wang2021real}.

In this paper we propose a novel method we call \textbf{Mixture of Diffusers}, that is able to run several diffusion processes simultaneously over the same canvas, all of them using the noise-predicting neural network, and harmonizing their contributions towards a shared image generation task. Figure \ref{fig:modTitle_MoD} presents an image sample generated through the use of three diffusion processes, each focusing on a different canvas region and using a different prompt from Figure \ref{fig:compositionPrompts}  (\textbf{Left}, \textbf{Center} and \textbf{Right}). The image shows a much better alignment to the intended composition than the results obtained with a single diffusion model.

The proposed Mixture of Diffusers method introduces several benefits:

\begin{itemize}
    \item Each diffusion model can act on a different region of the canvas, using a different prompt, thus allowing to \textbf{generate objects at specific locations} of the image, or introducing \textbf{smooth transitions in space between styles}.
    \item Since all diffusion processes share the same neural network, the memory footprint of the presented method is essentially the same as the memory required by the largest region affected by a single diffusion process. This allows \textbf{generating high resolution images in low-memory GPUs}.
    \item Mixture of Diffusers is compatible with SDEdit \cite{meng2021sdedit}, thus allowing conditioning the generation on images together with text prompts, thus allowing "image to image" and outpainting.
    \item The proposed method \textbf{can use any pre-trained diffusion model}. Even diffusion models that work in a latent space are amenable to be used, as long as they allow defining an approximate mapping between pixels in input space and locations in latent space.
\end{itemize}

\section{Background}
\label{sec:background}

\subsection{Diffusion models}

\begin{figure}
    \centering
    \includegraphics[width=0.9\textwidth]{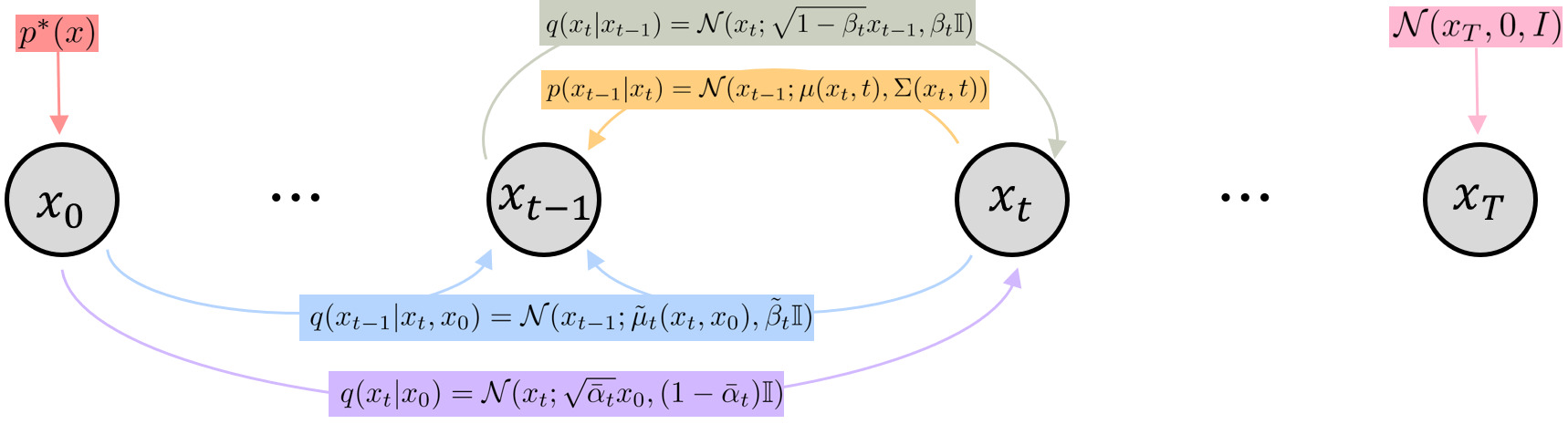}
    \caption{Illustration of the main equations governing a diffusion process.}
    \label{fig:diffusionEquations}
\end{figure}

\textbf{Diffusion models} \cite{sohl2015diffusionmodels} address the problem of obtaining samples from an unknown smooth probability distribution $p^*(x)$. Because $p^*(x)$ might be complex and might not be well approximated through a family of parameterized distributions, to make the problem more approachable a diffusion (noising) process is defined in the form of a Markov chain as follows:

\begin{equation}
    \label{eq:diffusionProcess}
    q(x_t|x_{t-1}) = \mathcal{N}(x_t; \sqrt{1 - \beta_t} x_{t-1}, \beta_t \mathbb{I}),
\end{equation}

where the expression $\mathcal{N}(z; \mu, \Sigma)$ must be understood as the probability of sampling $z$ from a gaussian distribution with mean $\mu$ and covariances matrix $\Sigma$. The process starts with $x_0$, a sample from the target distribution $p^*(x)$, and by iterating Equation \ref{eq:diffusionProcess} it produces gradually noisier versions $x_1, x_2, \ldots, x_T$ through the injection of gaussian noise. Such process has a number of convenient properties, which are summarized visually in Figure \ref{fig:diffusionEquations}; for starters, one can jump directly from $x_0$ to any $x_t$ also by introducing a different degree of gaussian noise,

\begin{equation}
    \label{eq:diffusionProcessTsteps}
    q(x_t|x_0) = \mathcal{N}(x_t; \sqrt{\bar{\alpha}_t} x_0, (1- \bar{\alpha}_t) \mathbb{I}),
\end{equation}

where the coefficients are now given by $\bar{\alpha}_t = \prod_{s=1}^t \alpha_s$, $\alpha_t = 1 - \beta_t$.

Related to this property, one can also "trace back" the diffusion process, obtaining $x_{t-1}$ from $x_t$, given the starting sample $x_0$ is known. Such transition  $q(x_{t-1}|x_t, x_0)$ once again follows a gaussian expression

\begin{equation}
    \label{eq:ddpmposterior}
    q(x_{t-1}|x_t, x_0) = \mathcal{N}(x_{t-1}; \tilde{\mu}_t(x_t, x_0), \tilde{\beta}_t \mathbb{I}),
\end{equation}

with means and variance scale given by

\begin{center}
\begin{tabular}{p{9cm}p{5cm}}
  \begin{equation}
    \label{eq:ddpmposteriormean}
    \tilde{\mu}_t(x_t, x_0) = \frac{\sqrt{{\bar \alpha}_{t-1}} \beta_t}{1 - {\bar \alpha}_t} x_0 + \frac{\sqrt{\alpha_t} (1 - {\bar \alpha}_{t-1})}{1 - {\bar \alpha}_t} x_t,
  \end{equation}
  &
  \begin{equation}
    \tilde{\beta}_t = \frac{1 - {\bar \alpha}_{t-1}}{1 - {\bar \alpha}_t} \beta_t.
  \end{equation}
\end{tabular}
\end{center}

Even more interesting is the fact that for sufficiently small noise variance values $\beta_t$ and large enough $T$ it can be shown that at the end of the process $x_T$ follows a standard gaussian distribution: $p(x_T) = \mathcal{N}(x_T, 0, I)$. Therefore, generating a sample from $x_T$ is trivial. Furthermore, a \textbf{reverse process} also exists, taking again the form of a Markov process with gaussian transitions

\begin{equation}
    \label{eq:reversediffusionProcess}
    p(x_{t-1}|x_t) = \mathcal{N}(x_{t-1}; \mu(x_t, t), \Sigma(x_t, t)),
\end{equation}

for unknown mean $\mu(x_t, t)$ and covariances $\Sigma(x_t, t)$ that depend on the process step $t$ and the current sample in the process $x_t$. 

Notably, these gaussian transition parameters can be modelled by using approximating functions (e.g. deep neural networks) in the form $\mu_{\theta}(x_t, t)$ and $\Sigma_{\theta}(x_t, t)$, for a set of learnable parameters $\theta$. Since we have a full probabilistic description of the process, if a dataset of samples $x_0$ from the target distribution is available, we can optimize such parameters $\theta$ so as to maximize the log-likelihood of the reverse process generating such samples when initialized with a sample from $x_T$,

\begin{equation}
    \max_{\theta} \mathbb{E}_{x_0 \sim p^*(x_0)}[\log p_{\theta}(x_0)] = 
    \max_{\theta} \mathbb{E}_{x_0 \sim p^*(x_0)} \left[ \log \left( p(x_T) \prod_{t=1}^{T} p_{\theta} (x_{t-1} | x_t)  \right) \right],
\end{equation}

with $p_{\theta} (x_{t-1} | x_t) = \mathcal{N}(x_{t-1}; \mu_{\theta}(x_t, t), \Sigma_{\theta}(x_t, t))$. To make this optimization problem tractable, one can follow the well-known procedure of optimizing the Evidence Lower Bound (ELBO) of the log-likelihood, in a fashion similar to the training of variational autoencoders \cite{kingma2013variationalautoencoders}:

\begin{equation}
    \max_{\theta} \mathbb{E}_{x_0 \sim p^*(x_0)}[\log p_{\theta}(x_0)]
    = \min_{\theta} \mathbb{E}_{x_0 \sim p^*(x_0)}[- \log p_{\theta}(x_0)] 
    \leq \min_{\theta} \mathbb{E}_{x_0 \sim p^*(x_0)} \left[ - \log \frac{p_{\theta}(x_{0:T})}{q(x_{1:T}|x_0)} \right] .
\end{equation}

Further derivation along this equation (we refer to Appendix A in \cite{ho2020denoising} for details) leads us to the objective function

\begin{equation}
    \label{eq:ELBOdiffusion}
    \max_{\theta} \mathbb{E}_{x_0 \sim p^*(x_0)} \left[ 
        D_{KL} \left( q(x_T|x_0) \; || \; p(x_T) \right)
        + \sum_{t>1}D_{KL} \left( q(x_{t-1}|x_t, x_0) \;|| \; p_{\theta}(x_{t-1}|x_t) \right)
        - \log p_{\theta}(x_0|x_1)
    \right] ,
\end{equation}

where $D_{KL}(z \; || \; w)$ is the Kullback-Leibler divergence between distributions $z$ and $w$. 

The expression in Equation \ref{eq:ELBOdiffusion} can be used to learn the values of the $\theta$ parameters, for instance by iteratively optimizing its terms through stochastic gradient descent, together with some variance reduction techniques \cite{ho2020denoising}. Regarding the values of the $\beta_t$ parameters, although they could also be possibly optimized with this method, they are generally set to a fixed schedule \cite{ho2020denoising, nichol2021improved}. Furthermore, the covariances can also be set to follow a fixed schedule such as $\Sigma_{\theta}(x_t, t) = \sigma_t^2 \mathbb(I)$, $\sigma_t^2 = \beta_t$ \cite{ho2020denoising}, or learned  by following a careful reparametrization to avoid training instabilities \cite{nichol2021improved}.

Once a model $(\mu_{\theta}(x_t, t), \Sigma_{\theta}(x_t, t))$ has been trained using the objective function above, new samples can be generated by starting from random noise $x_T \sim \mathcal{N}(x_T; 0, I)$ and iteratively applying the reverse process (Equation \ref{eq:reversediffusionProcess}) for each of the timesteps $\left[ T, \ldots, 1 \right]$ until arriving at a sample $x_0$.

\subsection{Learning a diffusion model by predicting noise}

As observed in \cite{ho2020denoising}, the learning strategy in diffusion models can be greatly simplified, resulting in an algorithm that is much simpler to implement and that even allows to produce better samples of $p^*(x)$. To perform this simplification, first note that Equation \ref{eq:diffusionProcessTsteps} can be reparametrized as

\begin{equation}
    \label{eq:xtfromeps}
    x_t(x_0, \epsilon) = \sqrt{\bar{\alpha}_t} x_0 + \sqrt{1 - \bar{\alpha}_t} \epsilon,
\end{equation}

for $\epsilon \sim \mathcal{N}(0, \mathbb{I})$. Which is to say, the sample $x_t$ at any step of the diffusion process can be obtained by mixing the initial sample $x_0 \sim p^*(x_0)$ with a sample from a standard gaussian distribution, using the schedule parameters as mixing weights. Following this expression, the learning objective could be telling apart the random noise $\epsilon$ from the actual data $x_0$ present in $x_t$, that is

\begin{equation}
    \label{eq:noiseLoss}
    L_{simple} := \mathbb{E}_{x_0 \sim p^*(x_0), t \sim [1, T], \epsilon \sim \mathcal{N}(0, \mathbb{I})} \left[ \left| \left| \epsilon - \epsilon_{\theta} (\sqrt{\bar{\alpha}_t} x_0 + \sqrt{1 - \bar{\alpha}_t} \epsilon, t) \right| \right|^2 \right],
\end{equation}

where $\epsilon_{\theta}(x_t, t)$ is a deep neural network that tries to predict the noise component present in $x_t$ at diffusion step $t$. This reformulation of the training objective has been shown to be a reweighing of the original terms derived by the ELBO above (Equation \ref{eq:ELBOdiffusion}) that puts less emphasis on the terms with smaller $t$ (less noise) \cite{ho2020denoising}. This approach is also closely related to score matching methods \cite{song2019generative, song2020score}.

The generation process using a model $\epsilon_{\theta}(x_t, t)$ is implemented by taking advantage of the fact that Equation \ref{eq:xtfromeps} can be rewritten as $x_0 = (x_t - \sqrt{1 - \bar{\alpha}_t} \epsilon_{\theta}(x_t)) / \sqrt{\bar{\alpha}_t}$, therefore we can get a (generally poor) estimate of the denoised data $x_0$ by using the current $x_t$ and noise estimated by the model. This estimate can then be used in Equation \ref{eq:ddpmposteriormean} to obtain the mean of $x_{t-1}$, which after some algebra (see Appendix \ref{app:posteriorMeanDerivation}) takes the form

\begin{equation}
    \tilde{\mu}_t(x_t, x_0) = \frac{1}{\sqrt{\alpha_t}} \left(
        x_t
        - \frac{1 - \alpha_t }{\sqrt{1 - {\bar \alpha}_t}} \epsilon_{\theta}(x_t, t) 
    \right) .
\end{equation}

Therefore a sample from $x_{t-1}$ is obtained through Equation \ref{eq:ddpmposterior} as

\begin{equation}
    \label{eq:ddpmepsgeneration}
    x_{t-1} = x_t
        - \frac{1 - \alpha_t }{\sqrt{1 - {\bar \alpha}_t}} \epsilon_{\theta}(x_t, t) + \frac{1 - {\bar \alpha}_{t-1}}{1 - {\bar \alpha}_t} \beta_t \; \mathcal{N}(0, \mathbb{I}) ,
\end{equation}

which can be seen as a combination of the current noisy data $x_t$, a step substracting the current estimation of the noise $\epsilon_{\theta}(x_t, t)$, and a random gaussian component. Thus, generating a new sample from $p^*$ involves starting once again from random noise $x_T \sim \mathcal{N}(x_T; 0, I)$, applying now Equation \ref{eq:ddpmepsgeneration} for each of the timesteps $\left[ T, \ldots, 1 \right]$ until we arrive at a sample $x_0$.

The sampling procedure described here is sometimes referred in the literature as the "DDPM scheduler" \cite{ho2020denoising}. Although effective, the procedure is slow, requiring $T$ iterations to transform a sample $x_T$ into a sample $x_0$. While it is possible to cut corners and accelerate this process by skipping steps, more refined methods or \textbf{schedulers} have been developed, such as DDIM \cite{song2020ddim} or LMS \cite{karras2022lms}, to name a few. Such schedulers mainly change the form of Equation \ref{eq:ddpmepsgeneration}, but still require obtaining predictions from a noise model $\epsilon_{\theta}(x_t, t)$. Thus, the choice of scheduler is an orthogonal decision to the way $\epsilon_{\theta}(x_t, t)$ is modelled. In the rest of the paper we focus on this last choice.


\subsection{Diffusion models for image generation}

While the diffusion techniques explained above can be potentially applied to any kind of target distribution $p^*(x)$, they have found prominent success in the field of image generation. Already in \cite{sohl2015diffusionmodels} the authors tested the generation of new samples by learning the data distribution of the popular MNIST \cite{lecun1998mnist} and CIFAR-10 \cite{krizhevsky2009cifar} image datasets. The reverse process is modelled by using a multi-scale convolutional network that outputs the means $\mu_{\theta, i}(x_t, t)$ and the variances $\sigma_{\theta,i} (x_t, t)$ for each pixel (off-diagonal terms of $\Sigma$ are set to $0$). A first section of the neural network reads $x_t$, generating two vectors of $J$ latent values for each pixel, each of them further used as an input to a linear function whose weights depend on the diffusion step $t$, producing $\mu_{\theta, i}(x_t, t)$ and $\sigma_{\theta,i} (x_t, t)$. In this way, a single network that is able to combine both image ($x_t$) and time ($t$) information is trained to run the denoising process.

Later on, \cite{ho2020denoising} address the higher resolution image datasets CelebA-HQ \cite{karras17celebqa} and LSUN \cite{yu2015lsun}, and propose the simplified learning objective of Equation \ref{eq:noiseLoss} focusing on predicting noise. For $\epsilon_{\theta}(x_t, t)$, a neural network architecture inspired by the U-Net \cite{ronneberger2015u} is also introduced, with the diffusion step $t$ being injected directly into each residual block in the form of a sinusoidal position embedding \cite{vaswani2017attention}.

A turning point in the image generation literature was established when in \cite{dhariwal2021diffusion} diffusion models were shown capable of producing better results than the state of the art GAN models, obtaining better Fréchet Inception Distance scores \cite{heusel2017fid} on several resolutions of the ImageNet dataset \cite{russakovsky2015imagenet}. Such results were attained through a careful design of the network architecture, presenting an architecture dubbed as Ablated Diffusion Model (ADM). Another innovation introduced in this work is the ability to condition the generation by means of an auxiliary classifier: suppose we have a classification model $p_{\phi}(y|x_t)$ that estimates the probability of data $x_t$ belonging to some particular class $y$. One can then define an inverse process that takes the form $p_{\theta, \phi}(x_{t-1}|x_t, y) = Z p_{\theta}(x_{t-1}|x_t) p_{\phi}(y | x_{t-1})$, with $Z$ a normalization constant. With this idea in mind, it is possible to derive update rules similar to Equation \ref{eq:ddpmepsgeneration} that take into account the output of the classifier $p_{\phi}(y | x_{t-1})$ to guide the generation process.

Alternatively, it is possible to embed  this guidance into the diffusion model itself: such strategy is known as classifier-free guidance \cite{ho2022classifierfreeguidance}. In essence, a neural network $\epsilon_{\theta}(x_t, t, y)$ is trained to predict the noise as shown in Equation \ref{eq:noiseLoss}, but the network also receives the class $y$ as additional information for this prediction. However, during the training procedure this class information is randomly dropped out, forcing the network to predict the noise without it. We note this as $\epsilon_{\theta}(x_t, t, \emptyset)$. At generation time the guided prediction for the noise is then

\begin{equation}
    \label{eq:classifierfreeguidance}
    {\hat \epsilon}_{\theta}(x_t, t, y) = \epsilon_{\theta}(x_t, t, \emptyset) + s \left( \epsilon_{\theta}(x_t, t, y) - \epsilon_{\theta}(x_t, t, \emptyset) \right) ,
\end{equation}

with $s$ a parameter that regulates the strength of the guidance. ${\hat \epsilon}_{\theta}(x_t, t, y)$ is used as a corrected noise estimate in the usual scheduler updating rules (e.g. Equation \ref{eq:ddpmepsgeneration}).

A clear advantage of classifier-free guidance was presented in the GLIDE method \cite{nichol2021glide}, where the class information $y$ is given as free-form text, also called prompt. The model $\epsilon_{\theta}(x_t, t, y)$ processes the text $y$ by encoding it as a sequence of tokens, then processing it through a stack of Transformer blocks to obtain a sequence of embeddings. Such embeddings are used in place of the class embedding of the ADM network, and also as extra tokens in the transformer layers of this network. As a result, the model is able to generate an image based on a text description.

A further refinement of these ideas was presented in the DALL-E 2 method \cite{ramesh2022hierarchical}, in which a multi-stage generation procedure was proved to produce results of better quality. A CLIP model \cite{radford2021clip} is used to transform the input prompt into a text embedding, which is then casted to an image embedding using a diffusion model referred to as a "diffusion prior". Making use of this embedding, a sequence of three diffusion models (also known as Cascaded Diffusion Model \cite{ho2022cascaded}) is used to produce a low resolution image ($64 \times 64$ pixels), and then upscale it up to the much higher resolution of $1024 \times 1024$ pixels. DALL-E 2 was shortly followed by Imagen \cite{saharia2022photorealistic}, an alternative approach making use of a pre-trained T5 language model \cite{raffel2020exploring} to obtain a sequence text embeddings, instead of the CLIP and diffusion prior models. This change in desing, together with a dynamic thresholding technique to stabilize classifier-free guidance, produce better quality results as judged by human observers.

Finally, another improvement in terms of efficiency is presented by Latent Diffusion Models (LDM) \cite{rombach2022high}. An autoencoder formed by the pair of networks $(E, D)$ is trained so as to compress a given image $x$ down to a latent space $E(x)$, in such a way that the original image can be recovered by applying $D(E(x)) \simeq x$. The diffusion model is then trained over the latent representation of images $E(x)$, together with a stack of Transformer layers fitted to produce text-embeddings of the input prompts. After training, the model can generate new images $z$ in the latent space, which are recovered as images in pixel space as $D(z)$. The popular Stable Diffusion \cite{stableDiffusion} model is essentially an LDM trained over a subset of the LAION-5B image dataset \cite{schuhmann2022laion}.

\subsection{Controlling composition in the generation process}

Some methods previous to this work have addressed the problem of controlling the location of objects or the composition of the image to be generated by the difussion model, largely inspired by previous successes of GAN-based methods. In SDEdit \cite{meng2021sdedit} the authors propose the usage of a guide image $x^{(g)}$ to initialize the denoising process at an intermediate step $t \in [0, T]$, which in the light of Equation \ref{eq:diffusionProcessTsteps} takes the form $x_t \sim \mathcal{N}(\sqrt{\bar{\alpha}_t} x^{(g)}, (1 - \bar{\alpha}_t) \mathbb{I})$. The choice of $t$ stands as a trade-off between realism ($t \simeq T$), understood as producing images in line with the training distribution $p^*(x)$, and faithfulness ($t \simeq 0$) in terms of similarity with the guide image $x^{(g)}$. This procedure is easy to implement, and has been popularized as the "img2img" algorithm. However, it requires the user to provide a guide image $x^{(g)}$, and the choice of $t$ might easily drift from following $x^{(g)}$ too closely, to largely ignoring it.

In the Palette method \cite{saharia2022palette} the authors address the tasks of image inpainting, uncropping (also known as outpainting), colorization and removal of JPEG compression artifacts. Inpainting and outpainting are relevant tools for controlling composition, as they allow to regenerate certain regions of the scene, or to extrapolate the canvas in any given direction. For inpainting, the authors train a diffusion model over images with regions of pixels removed from its center, by following a free-form or a rectangular mask. The pixel values in those regions are replaced by standard gaussian noise in $x_T$, and the loss function of the model is modified to consider only pixels in these regions. For outpainting the pixels at the borders at the image are removed, training a model in a similar fashion.

Following this trend, in GLIDE \cite{nichol2021glide} the authors perform a fine-tuning of their text-guided diffusion model so as to allow it to perform inpainting. The network receives four additional input channels, containing the RGB values of the reference image as well as a mask channel marking the region to inpaint. The fact that the diffusion model is conditioned on a prompt allows some degree of control over objects location, as the user can generate a first version of the image and then select a specific region to inpaint with a particular prompt referencing the object of interest. It must be noted, however, that the provided prompt should still describe the whole picture, as the diffusion model acts over the whole image.

A more fine-grained control over the composition was made possible with eDiff-I \cite{balaji2022eDiff-I}, where the user is allowed to select a region in the canvas and link it to a set of words in the text prompt, a method dubbed as {\it paint-with-words}. The attention layers of the model implementing $\epsilon_{\theta}(x_t, t, y)$ are modified to add an extra attention score between the token embeddings of the selected words and the pixels at the region of choice (appropriately downscaled to the working resolution of the attention layer). By defining several links between canvas regions and prompt words in this way, the generation process is biased to generate the objects in the regions of interest. Such method can be applied over an already trained model, just by modifying the architecture of the attention layers.

\section{Mixture of Diffusers}
\label{sec:mod}

\subsection{Method description}


We now introduce our method, Mixture of Diffusers, as a way to both control composition and to allow generating large images with a limited GPU RAM consumption. The key idea is to make use of $D$ noise prediction models $( {\hat \epsilon}_{\theta}^1, \ldots, {\hat \epsilon}_{\theta}^D)$ that build up the global noise prediction, each of them using a different text prompt $y_i$, and acting on a different (rectangular) region of pixels $x_{t,r_i}$, with $r_i = (a_i:b_i,c_i:d_i)$ an index of rows $[a_i,b_i)$ and columns $[c_i, d_i)$. In this way, the contribution of each model to the overall noise estimation takes a form similar to classifier-free guidance (Equation \ref{eq:classifierfreeguidance}),

\begin{equation}
    {\hat \epsilon}_{\theta}^i(x_{t,r_i}, t, y_i) = \epsilon_{\theta}^i(x_{t,r_i}, t, \emptyset) + s_i \left( \epsilon_{\theta}^i(x_{t,r_i}, t, y_i) - \epsilon_{\theta}^i(x_{t,r_i}, t, \emptyset) \right) .
\end{equation}

The overall noise prediction is then obtained through a weighted mixture of each individual noise prediction

\begin{equation}
    {\hat \epsilon}_{\theta}(x_t, t, y_{1:D}) = Z \odot \sum_{i=1}^D w_i \odot Padding_{r_i} \left( {\hat \epsilon}_{\theta}^i(x_{t,r_i}, t, y_i); x_t \right) ,
\end{equation}

where the $Padding_{r_i}(\hat x; x)$ operation generates a tensor with the same dimensionality as $x$, whose entries are those of $\hat x$ in the indices $r_i = (a_i:b_i, c_i:d_i)$, and $0$ elsewhere. The $w_i$ are tensors with the same dimensionality as $x_t$ containing mixing weights per pixel and model $i$, which get multiplied elementwise with the result of the padding operation. $Z$ is a normalization tensor such that $Z = \frac{1}{\sum_i w_i}$, where the division here must be understood as elementwise division; in this way the contribution of all models ${\hat \epsilon}_{\theta}^i(x_{t,r_i}, t, y_i)$ is normalized at the pixel level. The whole procedure is summed up in Algorithm \ref{alg:mod}.

\begin{algorithm}
\caption{Mixture of Diffusers}\label{alg:mod}
\begin{algorithmic}[1]
\Require Regions $r_{1, \ldots, D}$, prompts $y_{1, \ldots, D}$, guidance strengths $s_{1, \ldots, D}$, weights tensors $w_i$, diffusion steps $T$.
\State $x_T \sim {\cal N}(0, {\mathbb I})$  \Comment{Initialize from random noise}
\State $Z = \frac{1}{\sum_i w_i}$  \Comment{Compute weight normalization}
\For{$t \in [T, \ldots, 0]$}  \Comment{Denoising steps}
    \For{$i \in [1, \ldots, D]$}  \Comment{Compute noise predictions from each model}
        \State ${\hat \epsilon}_{\theta}^i(x_{t,r_i}, t, y_i) = \epsilon_{\theta}^i(x_{t,r_i}, t, \emptyset) + s_i \left( \epsilon_{\theta}^i(x_{t,r_i}, t, y_i) - \epsilon_{\theta}^i(x_{t,r_i}, t, \emptyset) \right)$
    \EndFor
    \State ${\hat \epsilon}_{\theta}(x_t, t, y_{1:D}) = Z \odot \sum_{i=1}^D w_i \odot Padding \left( {\hat \epsilon}_{\theta}^i(x_{t,r_i}, t, y_i); x_t, I_i \right)$ \Comment{Aggregate noise predictions}
    \State $x_{t-1} = Scheduler(x_t, {\hat \epsilon}_{\theta}(x_t, t, y_{1:D}))$ \Comment{Perform denoise step following scheduler}
\EndFor
\State \Return $x_0$
\end{algorithmic}
\end{algorithm}

In practice, for efficiency reasons, all models ${\hat \epsilon}_{\theta}^i$ use the same underlying noise-prediction network ${\hat \epsilon}_{\theta}$. Because of this, using more than one model does not necessarily result in a larger memory footprint, as only one model is loaded into memory and the model calls can be made sequentially. Alternatively, if ample memory is available, all calls to ${\hat \epsilon}_{\theta}$ can be batched together for better GPU efficiency.

\begin{figure}
    \centering
    \begin{subfigure}[b]{0.85\textwidth}
        \centering
        \includegraphics[width=\textwidth]{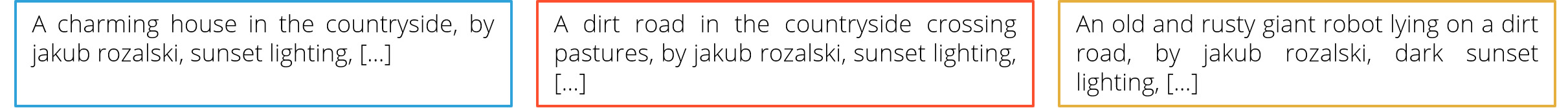}
    \end{subfigure}
    \begin{subfigure}[b]{0.85\textwidth}
        \centering
        \includegraphics[width=\textwidth]{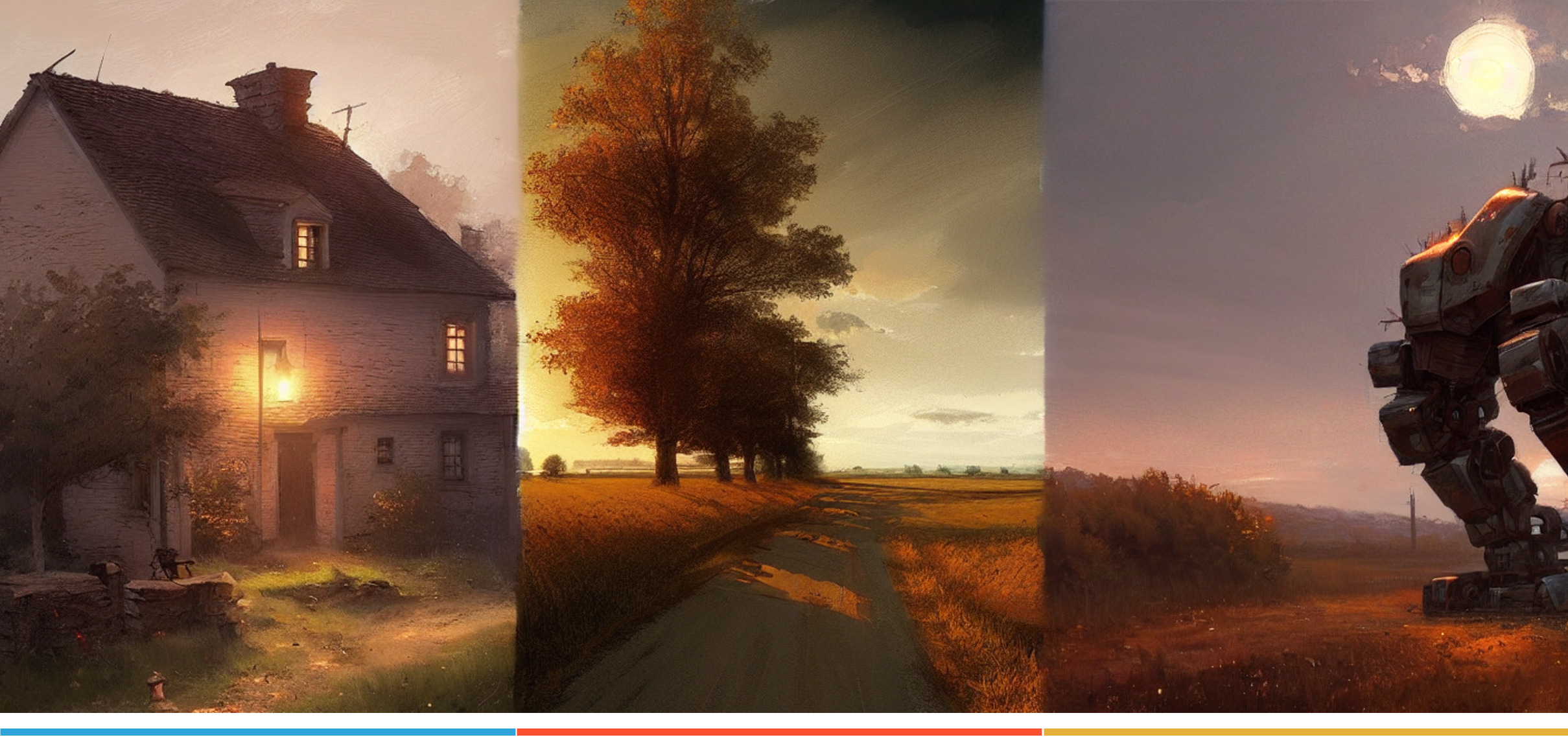}
    \end{subfigure}
    \begin{subfigure}[b]{0.85\textwidth}
        \centering
        \includegraphics[width=\textwidth]{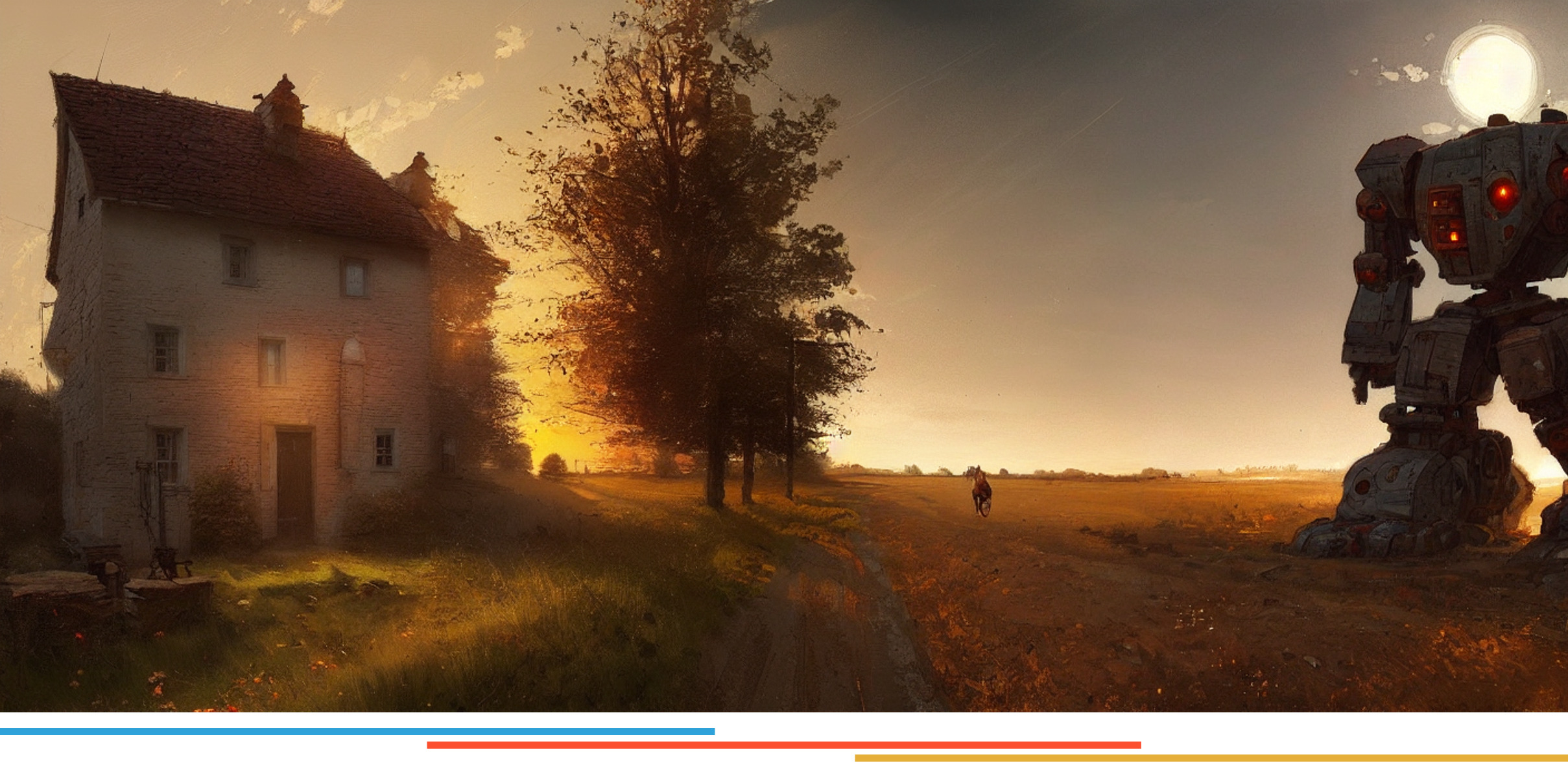}
    \end{subfigure}
    \begin{subfigure}[b]{0.85\textwidth}
        \includegraphics[width=1.02\textwidth, right]{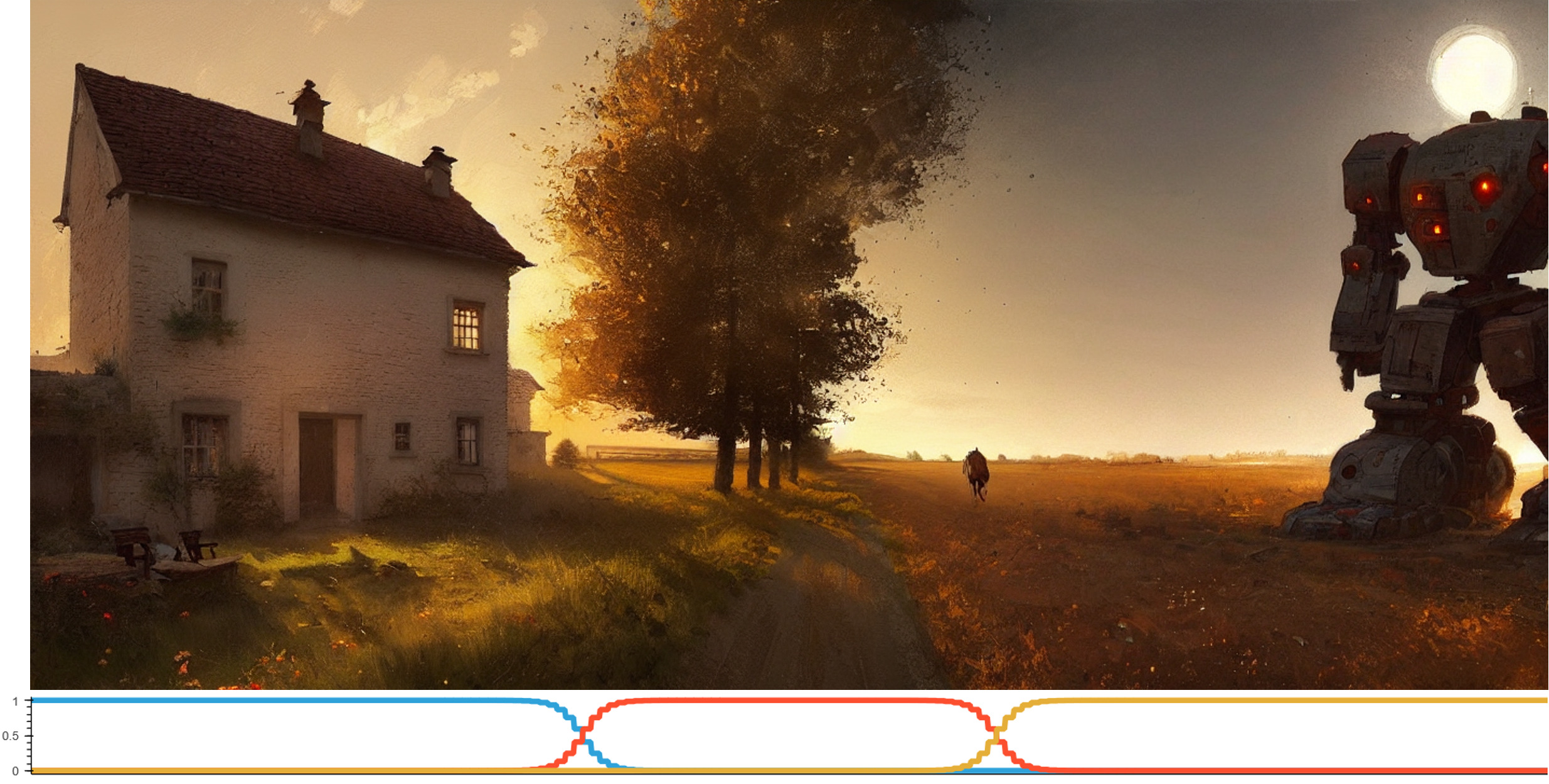}
    \end{subfigure}
    \caption{Effect of different region configuration strategies in Mixture of Diffusers. Each image has been generated using 3 models starting from the same initialization noise $x_T$ and prompts $y_i$, varying only the location of the regions affected by each model and their weights. {\bf Top}: each model operates on a different region, without overlapping. {\bf Center}: overlapping regions with constant weights. {\bf Bottom}: overlapping regions with gaussian weights.}
    \label{fig:modOverlapping}
\end{figure}

The usefulness of this formulation is in allowing the generation of a single image that is influenced by several noise-prediction models, each guided by a different prompt. In this way, it is easy to locate objects at specific regions of the image by adjusting the prompts correspondingly. As an illustrative example, Figure \ref{fig:modOverlapping} shows the result of creating an image using Stable Diffusion 1.4 as the noise-predicting model, together with the \textbf{Left} ($r_1$), \textbf{Center} ($r_2$) and \textbf{Right} ($r_3$) prompts of Figure \ref{fig:compositionPrompts}.

When each model acts on non-overlapping regions $r_i$ in the pixel space, the resultant image (Figure \ref{fig:modOverlapping} top) is essentially a concatenation of the outputs of three separate diffusion models, and the boundaries between each pair of models are clearly observed, as one might expect. This happens regardless of the choice of weights $w_i$.

However, a smooth transition between prompts can be created by overlapping regions $r_i$ between different models. For the center image in Figure \ref{fig:modOverlapping} we have configured each region $r_i$ to overlap 256 columns of pixels with each of the adjacent regions $r_{i-1}$ and $r_{i+1}$, using constant weights $w_i = 1$ (tensor full of ones). As a result we obtain a smooth transition between prompts, although some artifacts are still present: for instance, the right side of the house presents a marked vertical cut, exactly at the point where $r_2$ starts, and at the right side of this cut we can see tree branches with a floating window. The reason behind these artifacts is the usage of constant weights $w_i = 1$. Whenever a pixel is at the boundary of a region $r_i$ it will have neighbouring pixels that are no longer influenced by the corresponding model ${\hat \epsilon}_{\theta}^i$, and thus there will be a crisp difference in their overall noise predictions ${\hat \epsilon}_{\theta}$.

To solve this, we propose to use gaussian weights: for each pixel in $r_i$, its corresponding entry in the weights tensor $w_i$ will take the value of the probability density of a bidimensional gaussian with mean at the center of $r_i$. When measuring distance from any pixel to such center, distances are normalized by the width and height of $r_i$, and a value of $0.01$ is used for the gaussian variance. In this way, the weights take values close to $0$ when approaching the edges of $r_i$, favouring a smooth transition in pixel space between the influence of each model ${\hat \epsilon}_{\theta}^i$. Weights corresponding to pixels outside the region for $r_i$ take the value $0$. The bottom of Figure \ref{fig:modOverlapping} shows the resultant image of applying this strategy, together with a plot of the normalized weights ${\bar w}_i = Z \odot w_i$ at each column of the image (because of the normalization and how the regions are configured horizontally, all pixels in the same column share the same ${\bar w}_i$).

\subsection{Adaption to latent spaces}

While the method described above can be applied to any diffusion model that works in pixel space, in Latent Diffusion Models \cite{rombach2022high, stableDiffusion} the operation is performed over an image compressed into a latent space. More specifically, the generation process starts from a random sample $z_T$ in latent space, and the usual denoising steps are carried out to produce a sequence $z_{T-1}, \ldots, z_0$, finally obtaining an image in pixel space by using a decoder, $x_0 = D(z_0)$. Since Mixture of Diffusers is configured as a number of noise-predicting models on specific regions of pixel space, an adaption to latent space is required.

Ideally, we would like to have a mapping function able to translate diffusion regions $r_i$ defined as indices in pixel space to regions $r^l_i$ in latent space. It should be noted, however, that for a general decoder function $D$ a single position $z_{ij}$ in latent space might influence all the values in pixel space, thus resulting in a degenerate setting where each model ${\hat \epsilon}_{\theta}^i(x_{t,r_i}, t, y_i)$ must process the whole latent space, hence rendering the proposed method ineffective.

\begin{figure}
    \centering
    \includegraphics[width=0.5\textwidth]{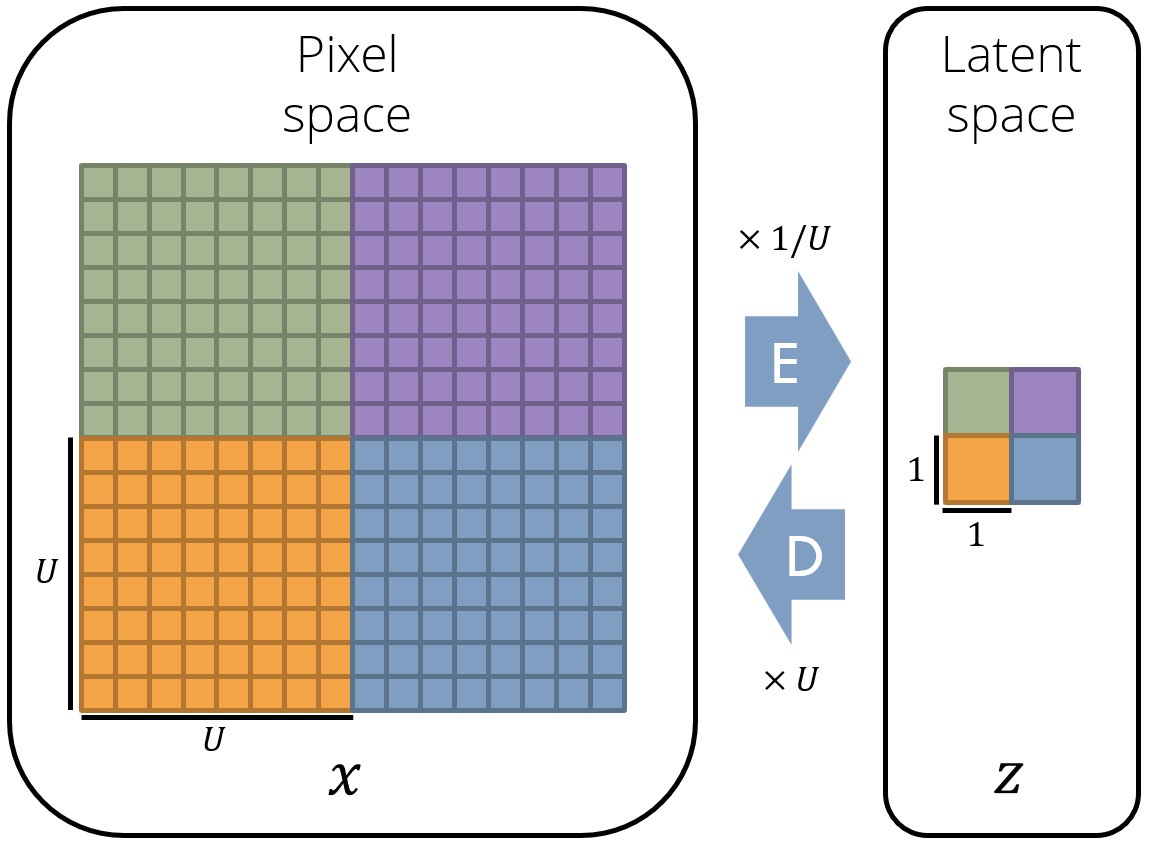}
    \caption{Mapping between pixel space and latent space used in Mixture of Diffusers. Each region of pixels specified by the user is mapped to a latent region of size divided by the upscaling factor $U$ of the encoder.}
    \label{fig:pixelLatentMapping}
\end{figure}

We propose to use an approximate but simple and effective mapping, which assumes that each location in latent space gives rise to a $U \times U$ region of values in pixel space, with $U$ the upscaling factor of the decoder $D$. Thus, when the user defines a rectangular region of the canvas in pixel space, $r_i = (a_i:b_i,c_i:d_i)$, we operate the corresponding noise-prediction model ${\hat \epsilon}_{\theta}^i(x_{t,r_i}, t, y_i)$ in the region $r^l_i = z_{(a_i/U):(b_i/U),(c_i/U):(d_i/U)}$, which is to say, we simply divide the index values by $U$. To avoid fractional indices, we require the canvas indices $(a_i, b_i, c_i, d_i)$ to be a multiple of $U$. More intuition on this idea is given in Figure \ref{fig:pixelLatentMapping}.

Although, in the following sections the experiments will show that this mapping works well in practice, we also provide here some intuition on why it is a good approximation. In Latent Diffusion Models the decoder is implemented by a convolutional neural network akin to the decoding half of a U-Net: a series of blocks mostly made of convolution, activation and normalization layers, with each block also including a single upsampling operation. For instance, Stable Diffusion \cite{stableDiffusion} uses 3 of such blocks with a $\times 2$ upsampling operation in each, thus producing a total upscaling factor of $U=8$. Critically, because of the local nature of convolutions, for large enough regions $r^l_i$ in latent space most of the pixels in the corresponding region $r_i$ in pixel space take values that indeed depend only on values contained in $r^l_i$, except for those pixels near the borders of $r_i$ whose values are also affected by latents outside of $r^l_i$. But since in the proposed method the pixels at the borders of $r_i$ are either at the edge of the canvas or are also covered by another overlapping region $r_j$, this problem is smoothed out.

We note also that most modern implementations of U-Nets include self-attention layers in some blocks, through which any value in latent space could potentially influence any other value in pixel space; in fact Stable Diffusion includes such a self-attention layer in the decoder before the first upsampling block. Our proposed approximate mapping ignores this fact. However, as the experiments will shown, this does seem not pose a problem in practice.

\section{Experiments}
\label{sec:experiments}

We now present various experiments illustrating the capabilities of Mixture of Diffusers. Our method has been implemented as an open-source library available at \url{https://github.com/albarji/mixture-of-diffusers}, which extends the popular Diffusers \cite{von-platen-etal-2022-diffusers} library from Hugging Face. All the results presented here have been produced with this implementation.

In all experiments the base model used was Stable Diffusion 1.4 \cite{rombach2022high, stableDiffusion}. We also use a Tesla P100 GPU with 16GB of RAM, to illustrate how the method can be applied with resources of capabilities akin to consumer hardware.

Additionally, all the experiments use the gaussian weights strategy described in the previous section.

\subsection{Image composition}

On top of the cherry-picked results presented in previous sections, Appendix \ref{app:composition} presents random samples (e.g. not selected or filtered in any way) from Stable Diffusion 1.4 and Mixture of Diffusers. The captions used for Stable Diffusion try to express the location where objects must be placed, while for Mixture of Diffusers this objective is attained by using several models with different captions and different regions.

In general, Stable Diffusion struggles to generate an image with all the objects described in the prompt. This is aligned with a general observation among Stable Diffusion users, by which the first tokens in the prompt tend to have a more significant imprint on the generated image. This effect hinders the user ability to employ the prompt as a way to present a detailed description of the desired composition.

It is also noticeable that Stable Diffusion also tends to generate samples that seem to be split into several images. This again might be due to a poor interpretation of the prompt, where specifying the location of objects guide the model to position several independent images in the canvas.

Contrary to this, Mixture of Diffusers correctly presents and locates the specified objects in almost all samples, although incoherencies appear in some samples, something especially noticeable when looking at the horizon and how it can change height across different sections of the image. This effect tends to happen when a large object occludes the horizon and each side of such object lies in the region of a different model.

\subsection{High resolution image generation}

Mixture of Diffusers has a constant memory cost, upper-bounded by the memory requirements of the most demanding ${\hat \epsilon}_{\theta}^i(x_{t,r_i}, t, y_i)$ in usage. In turn, the memory requirements of such model are proportional to the number of pixels in the region it operates ($x_{t,r_i}$). Thus it is possible to generate very large images by using a sufficiently large number of models with their corresponding regions.

\begin{figure}
    \centering
    \begin{subfigure}[b]{1\textwidth}
        \centering
        \includegraphics[width=\textwidth]{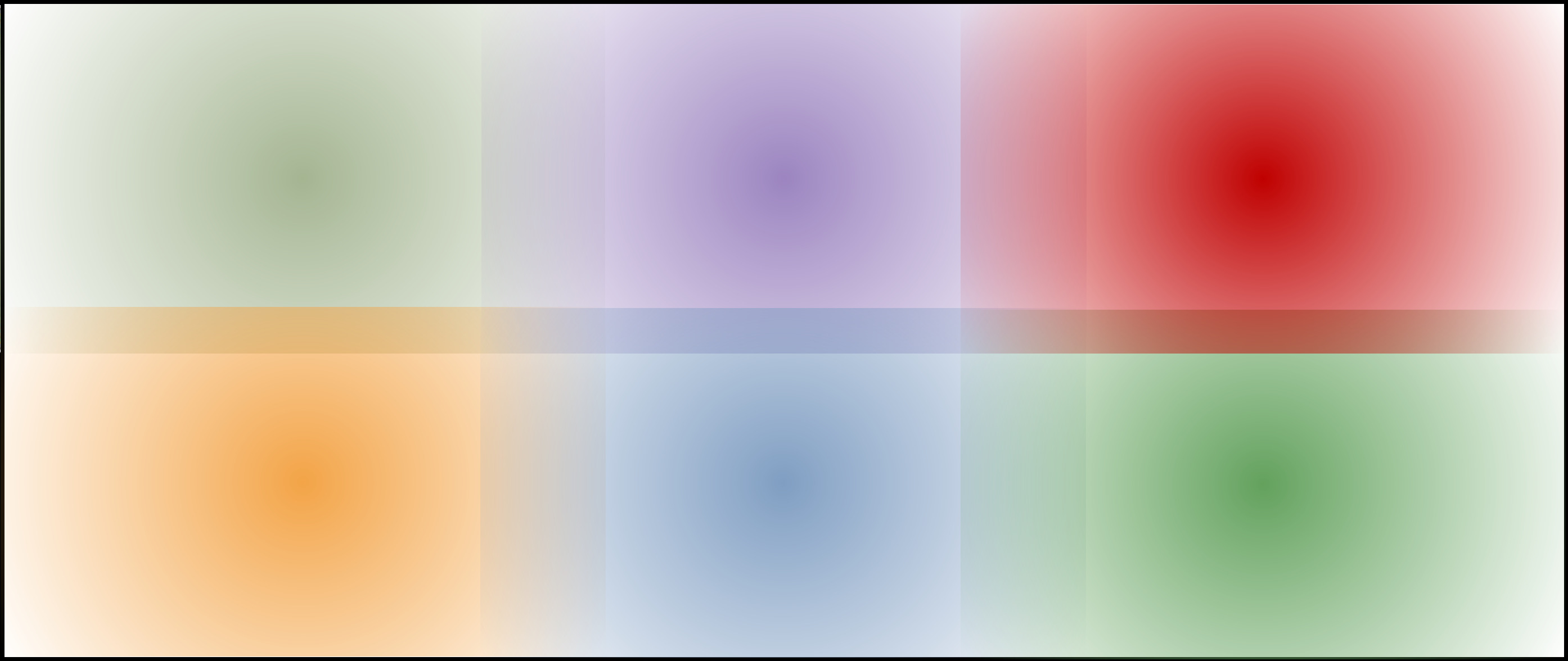}
    \end{subfigure}
    \begin{subfigure}[b]{1\textwidth}
        \centering
        \includegraphics[width=\textwidth]{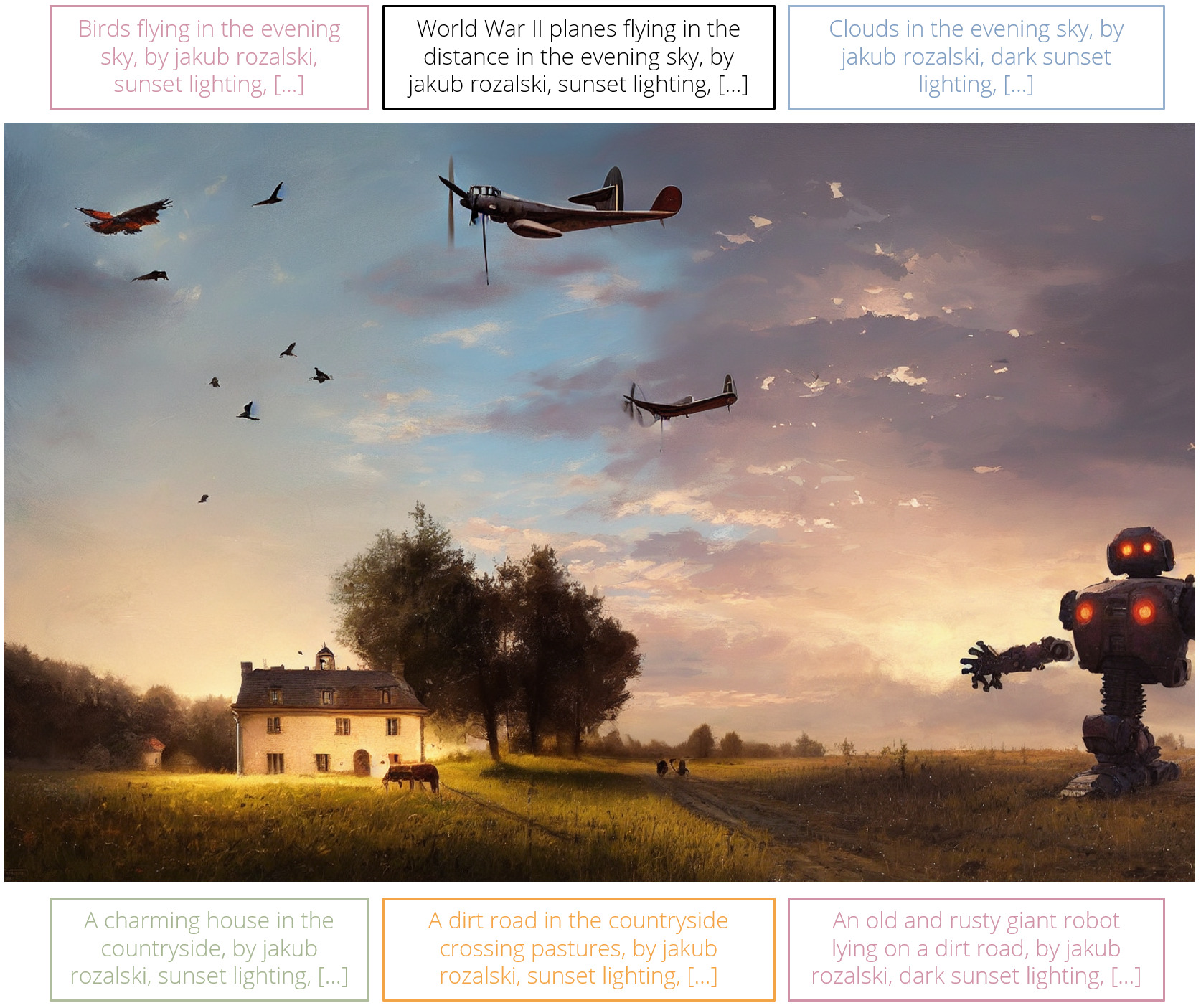}
    \end{subfigure}
    \caption{Generation of a 1408x896 image using 6 regions. {\bf Top}: spatial configuration of the 6 regions $r_i$. The shadings in each region represents the gaussian weights $w_i$. {\bf Bottom}: prompts and generated image. All prompts end with the style text "elegant, highly detailed, smooth, sharp focus, artstation, stunning masterpiece".}
    \label{fig:6regions}
\end{figure}

As an example, Figure \ref{fig:6regions} presents a 1408x896 image generated by making use of 6 regions arranged in an overlapping 2 rows x 3 columns grid, all of them employing the same Stable Diffusion 1.4 model. Each region covers an area of 640x640 pixels, with overlaps of 256 pixels in the column dimension and 384 in the row dimension. Generating an image of this size is not feasible using a single Stable Diffusion model with the hardware at hand.

\begin{figure}
    \centering
    \begin{subfigure}[b]{1\textwidth}
        \centering
        \includegraphics[width=\textwidth]{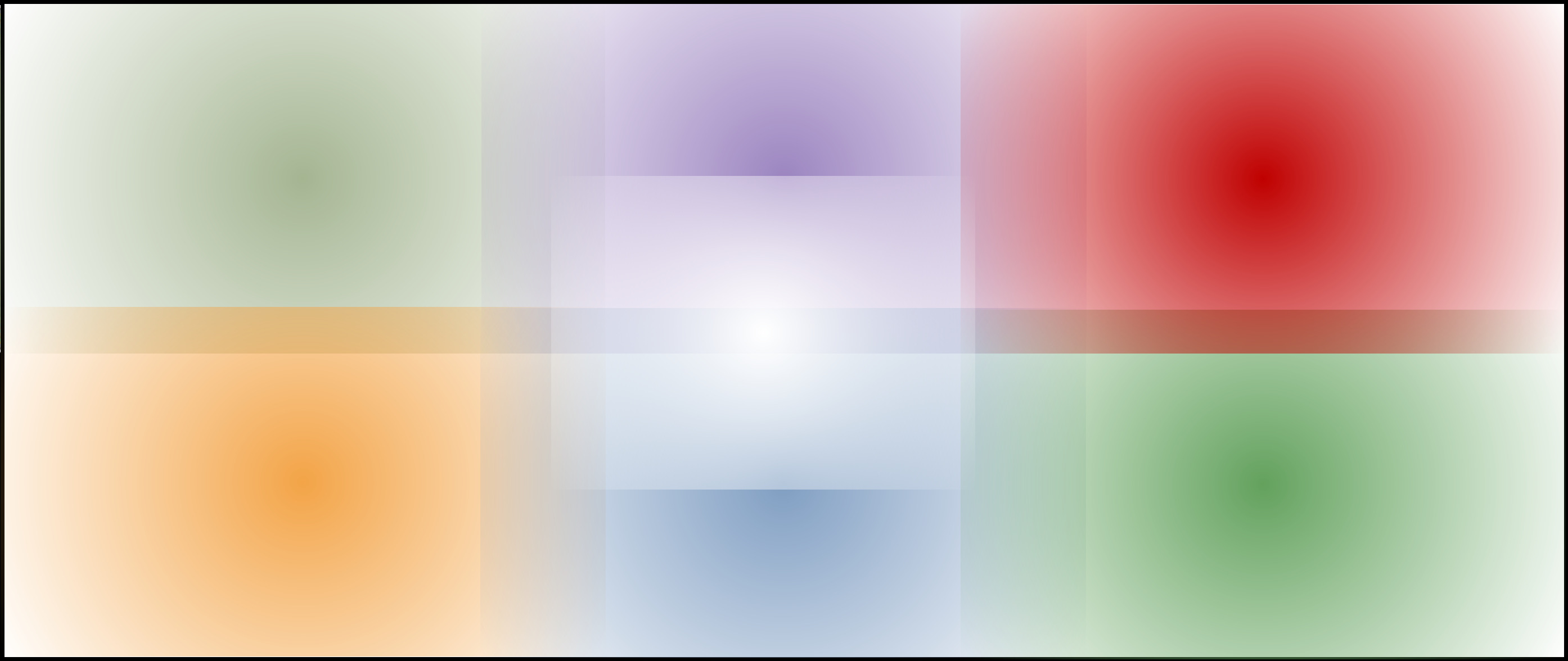}
    \end{subfigure}
    \begin{subfigure}[b]{1\textwidth}
        \centering
        \includegraphics[width=\textwidth]{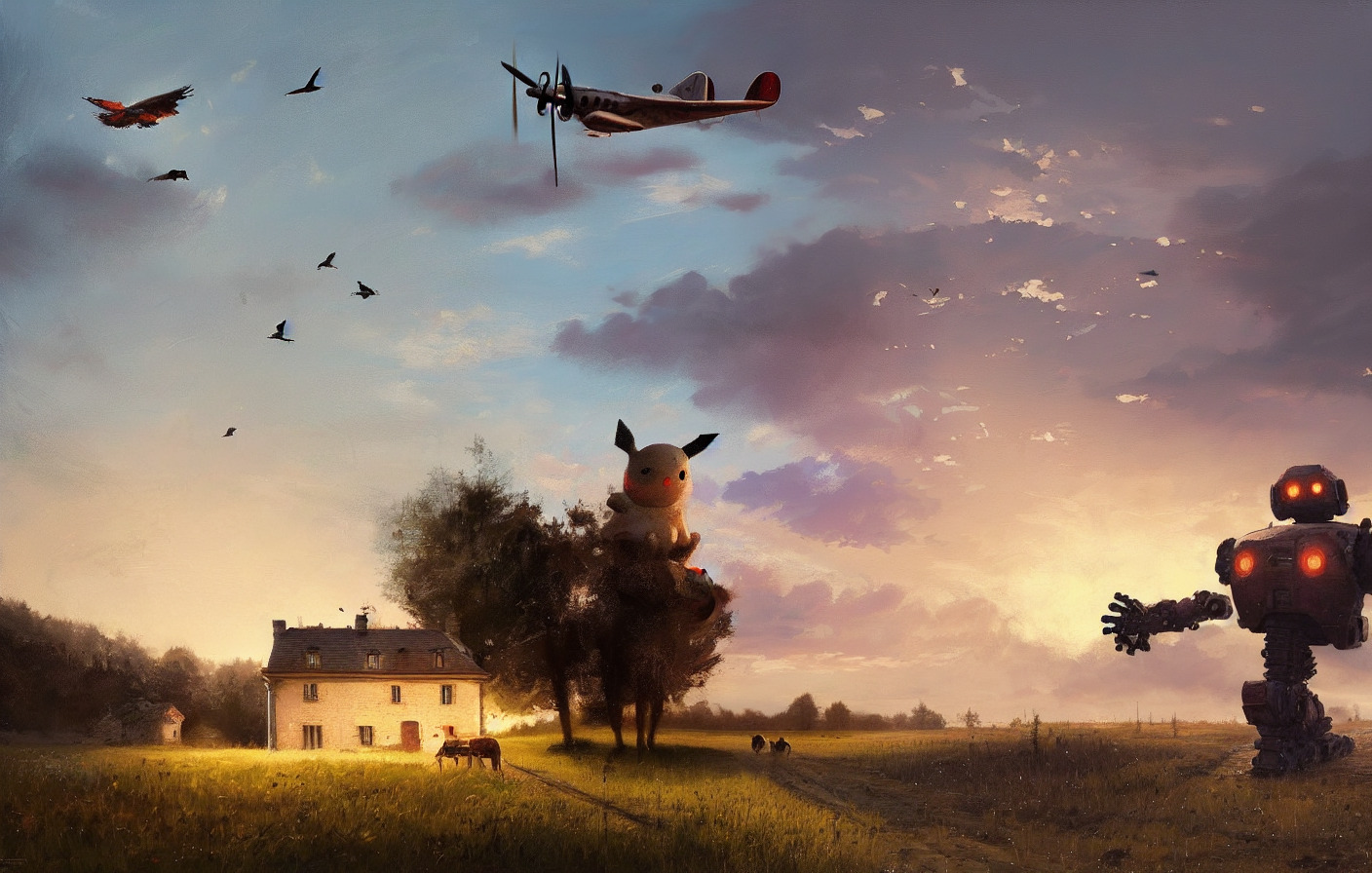}
    \end{subfigure}
    \caption{Extension of Figure \ref{fig:6regions} with an additional model acting in the center of the image. {\bf Top}: spatial configuration of the 7 regions $r_i$. The new region is presented as a white shading in the center. {\bf Bottom}: generated image. The prompt for the new region is "A giant pikachu, by jakub rozalski, dark sunset lighting, elegant, highly detailed, smooth, sharp focus, artstation, stunning masterpiece".}
    \label{fig:7regions}
\end{figure}

It is also worth noting that the proposed method is flexible enough to allow the placement of regions in any way, as long as they take a rectangular shape. For instance, the image already presented in Figure \ref{fig:6regions} can be further modified to add a new model acting on the center of the image, thus producing the image in Figure \ref{fig:7regions}. This allows for a significant degree of control in the placement of specific objects.

\begin{figure}
    \centering
    \includegraphics[width=\textwidth]{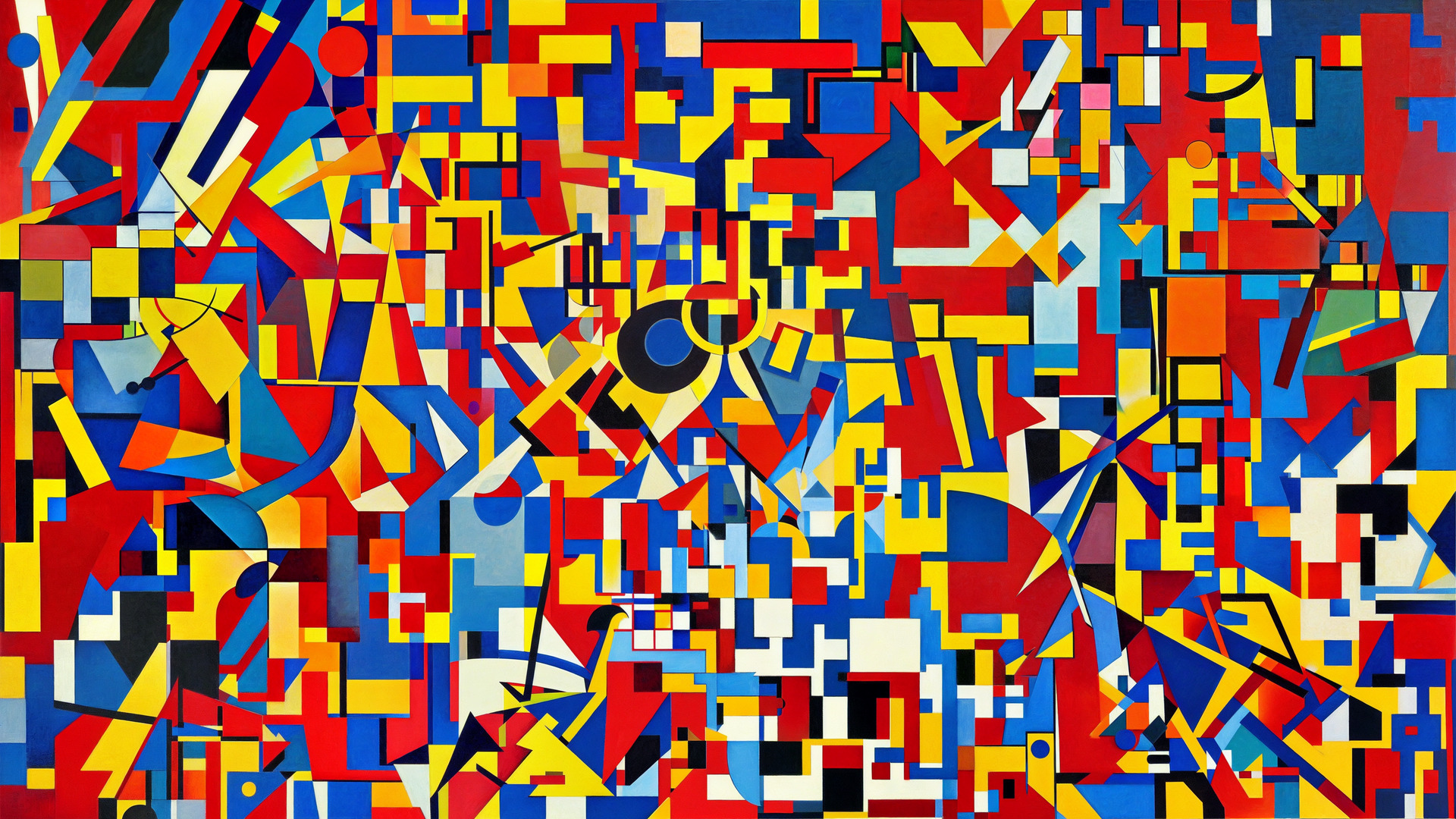}
    \caption{4K image generated with 8x11 models arranged in a grid, all of them using the prompt "Abstract decorative illustration, by lyubov popova and kadinski and kazimir malevich and mondrian, elegant, intricate, highly detailed, smooth, sharp focus, vibrant colors, artstation, stunning masterpiece". Full resolution image available at \url{https://albarji-mixture-of-diffusers-paper.s3.eu-west-1.amazonaws.com/4Kpopova.png}.}
    \label{fig:4kpopova}
\end{figure}

In our experiments we were able to extend this technique for high-resolution image generation up to 4K images (3840x2160) while still using a 16GB RAM GPU. Figure \ref{fig:4kpopova} shows an example where 88 models for regions of size 640x480 are arranged in a grid of 8 rows by 11 columns, with an overlap of 320 pixels with neighbouring top and bottom regions, and 240 pixels with neighbouring left and right regions. All models use the same prompt, "Abstract decorative illustration, by lyubov popova and kadinski and kazimir malevich and mondrian, elegant, intricate, highly detailed, smooth, sharp focus, vibrant colors, artstation, stunning masterpiece". Additional results are presented in Appendix \ref{app:4k}.

We should also note that when using a Latent Diffusion Model (such as Stable Diffusion in our experiments) the encoder and decoder networks do have a memory cost that grows linearly with the number of pixels in the images being processed. Still, since these networks are reasonable fast to evaluate, we can run the encoding and decoding processes in CPU, were RAM is cheaper resource.

\subsection{Smooth style transitioning}

\begin{figure}
    \centering
    \includegraphics[width=\textwidth]{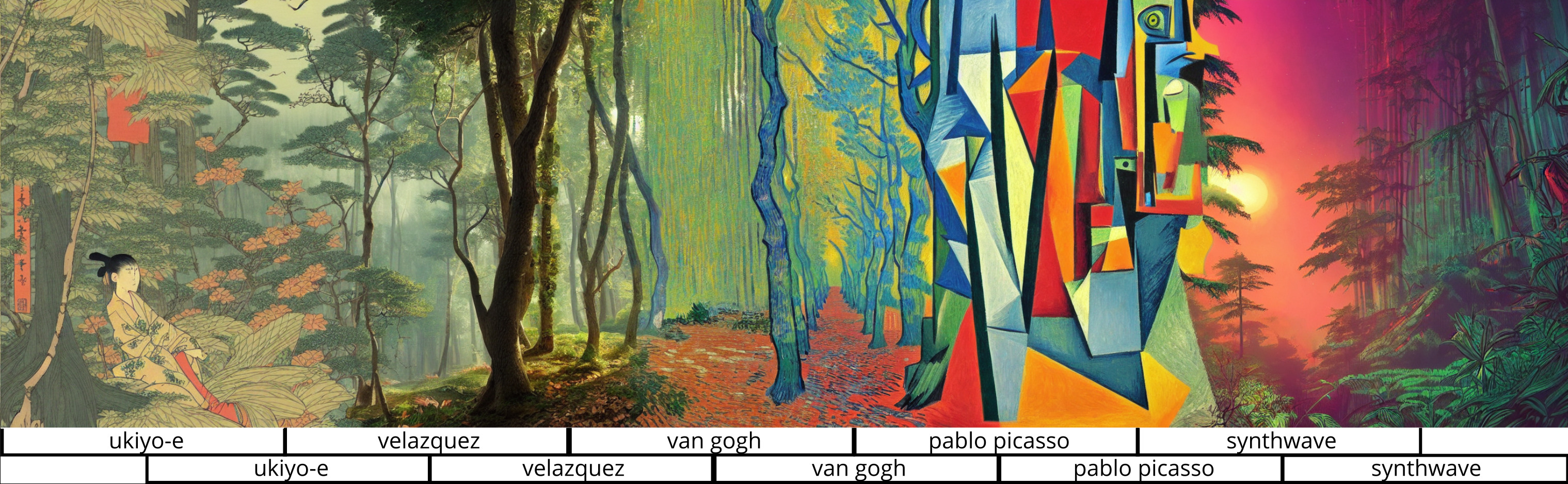}
    \caption{Example of smooth transitioning between different styles by making use of several regions. All regions use the prompt pattern "A forest, {\bf [STYLE]}, intricate, elegant, highly detailed, smooth, sharp focus, artstation, stunning masterpiece, impressive colors", where the {\bf [STYLE]} of each region is stated under the image.}
    \label{fig:linearforest}
\end{figure}

The ability to assign different prompts to each region of the canvas can also be used to generate smooth transitions between styles, by simply using the same base prompt in all regions, only varying the description of the styles to be used. Figure \ref{fig:linearforest} shows an example of a linear transition between 5 different styles. All prompts follow the pattern "A forest, {\bf [STYLE]}, intricate, elegant, highly detailed, smooth, sharp focus, artstation, stunning masterpiece, impressive colors". All regions have shape 768x512, with an overlap of 256. Two regions are used per style to allow for a slower transition and a clearer expression of each style.

\subsection{Conditioning on images}

A simple but useful extension that can be included in the presented method is the ability to condition a region of the canvas on a particular guide image. To do so, we adapt the strategy of SDEdit \cite{meng2021sdedit} in the following way: the user is able to provide one or more guide images $x^{(g_i)}$ together with placement regions $I^{(g_i)}$ (in the form of indices) and guidance strengths $t^{(g_i)}$. In SDEdit a single guide image influences the whole canvas, and thus the denoising process can be started at a intermediate step $x_t \sim \mathcal{N}(\sqrt{\bar{\alpha}_t} x^{(g)}, (1 - \bar{\alpha}_t) \mathbb{I})$ conditioned by the guide image $x^{(g)}$. However, in our proposed method we might find canvas regions influenced by one, several, or no guide image at all, a fact that depends on the placement indices $I^{(g_i)}$ of such images. Hence, we need a more general approach to image conditioning.

Such approach is actually simple, and is illustrated in Algorithm \ref{alg:conditioning}. At each step $t$ of the denoising process, the strength $t^{(g_i)}$ of each guidance image is checked, and if the guidance is still in effect ($t >= t^{(g_i)}$) the current pixel values at $x_{t-1}(I^{(g_i)})$ are replaced by a noisy version of $x^{(g_i)}$ at the current level of noise $t$. Since the same guide image will be processed in this way at different levels of noise $t$, to ensure a consistent noising pattern we use a deterministic noising scheduled based on the initial noise $x_T$

\begin{equation*}
    x_{t-1}(I^{(g_i)}) = \sqrt{\bar{\alpha}_t} x^{(g_i)} + (1 - \bar{\alpha}_t) x_T(I^{(g_i)}) .
\end{equation*}

\begin{algorithm}
\caption{Mixture of Diffusers {\color{blue} with image conditioning}}\label{alg:conditioning}
\begin{algorithmic}[1]
\Require Regions $r_{1, \ldots, D}$, prompts $y_{1, \ldots, D}$, guidance strengths $s_{1, \ldots, D}$, weights tensors $w_i$, diffusion steps $T$, {\color{blue} guide images $x^{(g)_{1, \ldots, G}}$, guide image placement regions $I^{(g_{1, \ldots, G})}$, guide image strengths $t^{g_{1, \ldots, G}}$}.
\State $x_T \sim {\cal N}(0, {\mathbb I})$  \Comment{Initialize from random noise}
\State $Z = \frac{1}{\sum_i w_i}$  \Comment{Compute weight normalization}
\For{$t \in [T, \ldots, 0]$}  \Comment{Denoising steps}
    \For{$i \in [1, \ldots, D]$}  \Comment{Compute noise predictions from each model}
        \State ${\hat \epsilon}_{\theta}^i(x_{t,r_i}, t, y_i) = \epsilon_{\theta}^i(x_{t,r_i}, t, \emptyset) + s_i \left( \epsilon_{\theta}^i(x_{t,r_i}, t, y_i) - \epsilon_{\theta}^i(r_{i,t}, t, \emptyset) \right)$
    \EndFor
    \State ${\hat \epsilon}_{\theta}(x_t, t, y_{1:D}) = Z \odot \sum_{i=1}^D w_i \odot Padding \left( {\hat \epsilon}_{\theta}^i(x_{t,r_i}, t, y_i); x_t, I_i \right)$ \Comment{Aggregate noise predictions}
    \State $x_{t-1} = Scheduler(x_t, {\hat \epsilon}_{\theta}(x_t, t, y_{1:D}))$ \Comment{Perform denoise step following scheduler}
    \For{{\color{blue} $i \in [1, \ldots, G]$}}  \Comment{{\color{blue} Process guidance images}}
        \If{{\color{blue}$t >= t^{g_i}$}}
            \State {\color{blue} $x_{t-1}(I^{(g_i)}) = \sqrt{\bar{\alpha}_t} x^{(g_i)} + (1 - \bar{\alpha}_t) x_T(I^{(g_i)})$}  \Comment{{\color{blue} Override with noisy version of guidance image}}
        \EndIf
    \EndFor
\EndFor
\State \Return $x_0$
\end{algorithmic}
\end{algorithm}

\begin{figure}
    \centering
    \includegraphics[width=\textwidth]{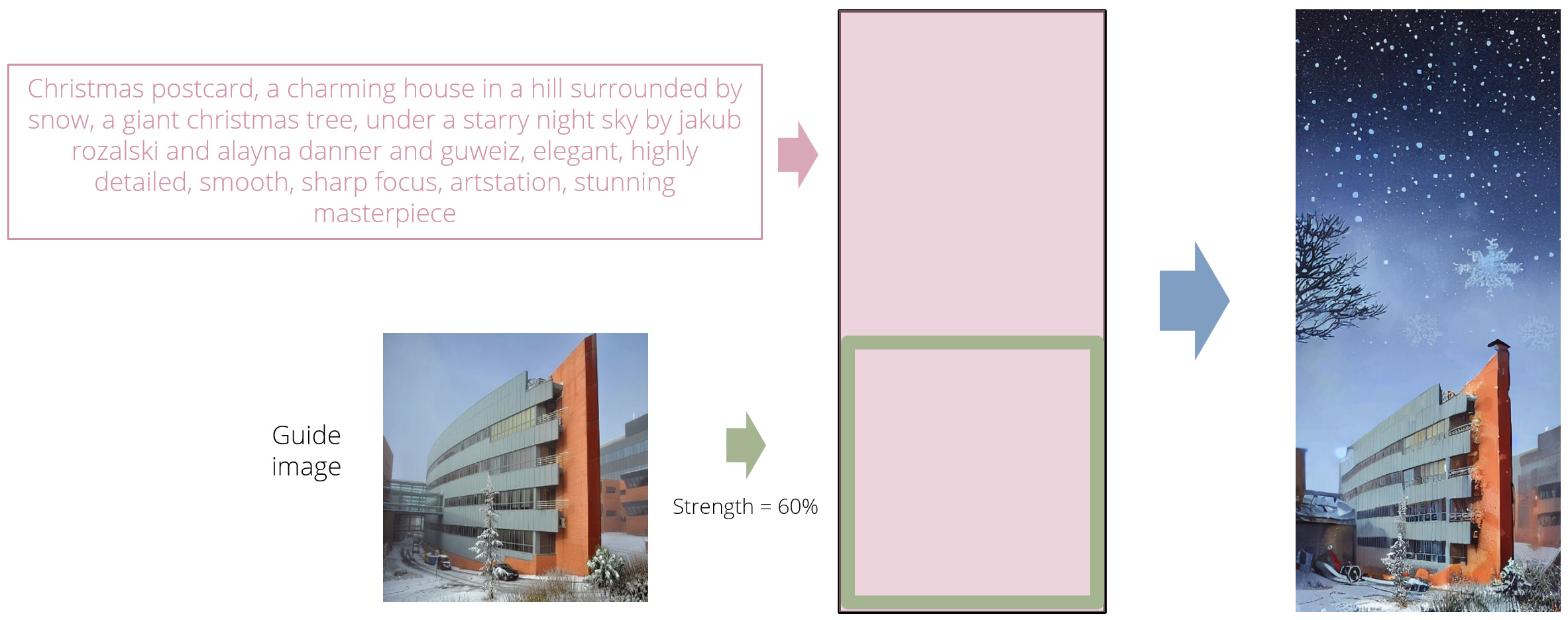}
    \caption{Example of Mixture of Diffusers with image conditioning. A guide image is used at the bottom of the canvas, with a strength of $60\%$ ($0.6T$), while a diffusion model guides the whole canvas toward a prompt.}
    \label{fig:imageconditioningIIC}
\end{figure}

An example of this kind of image conditioning is given in Figure \ref{fig:imageconditioningIIC}, where a guide image is used to condition part of the canvas, and a diffusion model guides the whole picture towards a prompt, in a form of outpainting, resulting in an overall consistent image.

\begin{figure}
    \centering
    \includegraphics[width=\textwidth]{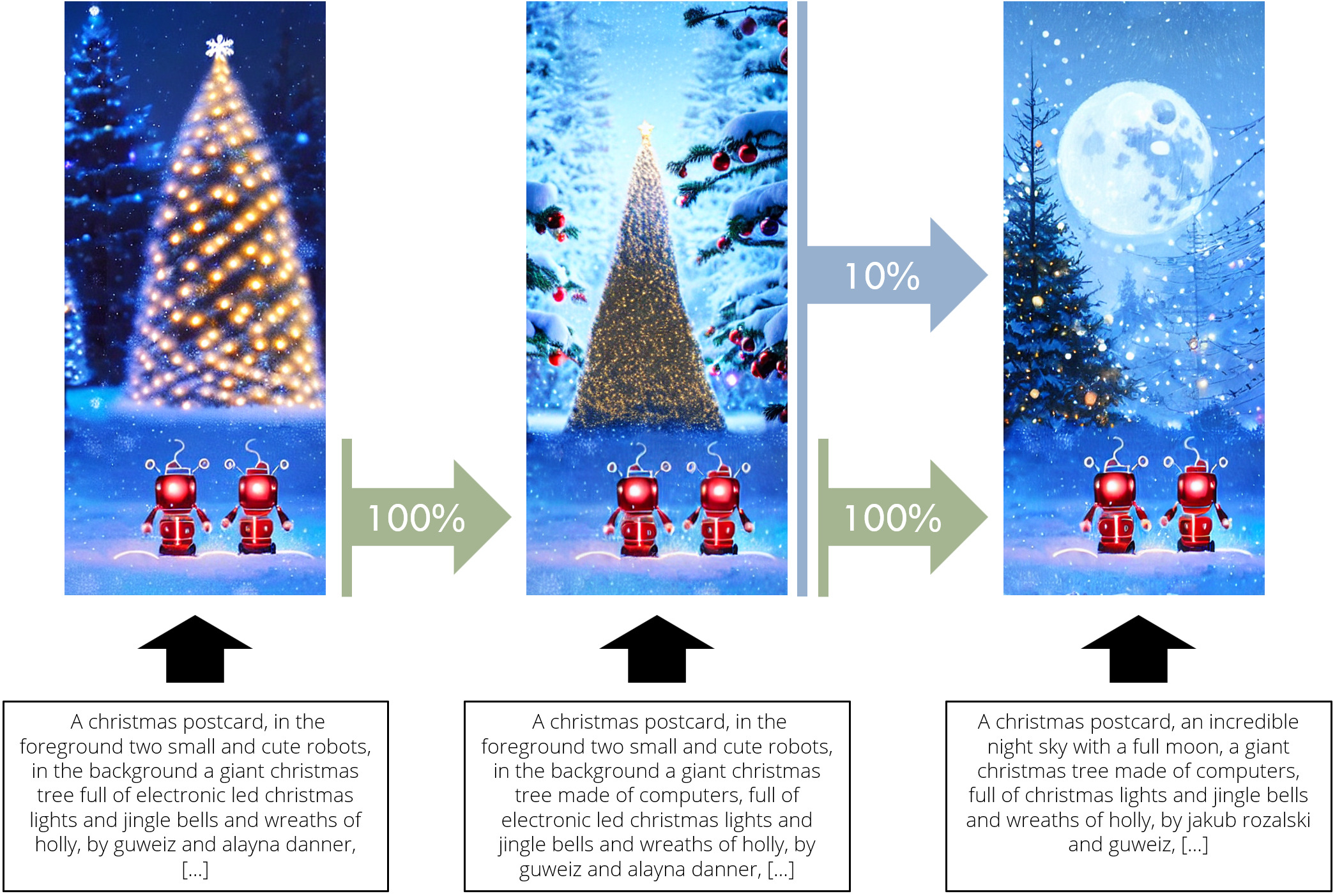}
    \caption{An example of an incremental workflow. As a first step an image is generated using only a text prompt. Then, the bottom part is fixed (guide strength $100\%$) and a new image is generated with a different prompt and initial noise. Finally, in the third version the bottom part is still fixed while the whole second image is used as guidance with $10\%$ strength in order to slightly preserve the general composition. In all prompts the tailing text is "elegant, highly detailed, smooth, sharp focus, artstation, stunning masterpiece".}
    \label{fig:incremental}
\end{figure}

The ability to use a given image as a guide for a part of the canvas allows for an iterative creation process, where part of a previously generated image can be fixed while generating a different version of the rest of the canvas, by means resampling the initial noise $x_T$ or modifying the prompts. An example of such an iterative is given in Figure \ref{fig:incremental}, where a postcard is generated in three steps.

Further outpainting experiments are presented in Appendix \ref{app:outpainting}.

\begin{figure}
    \centering
    \includegraphics[width=\textwidth]{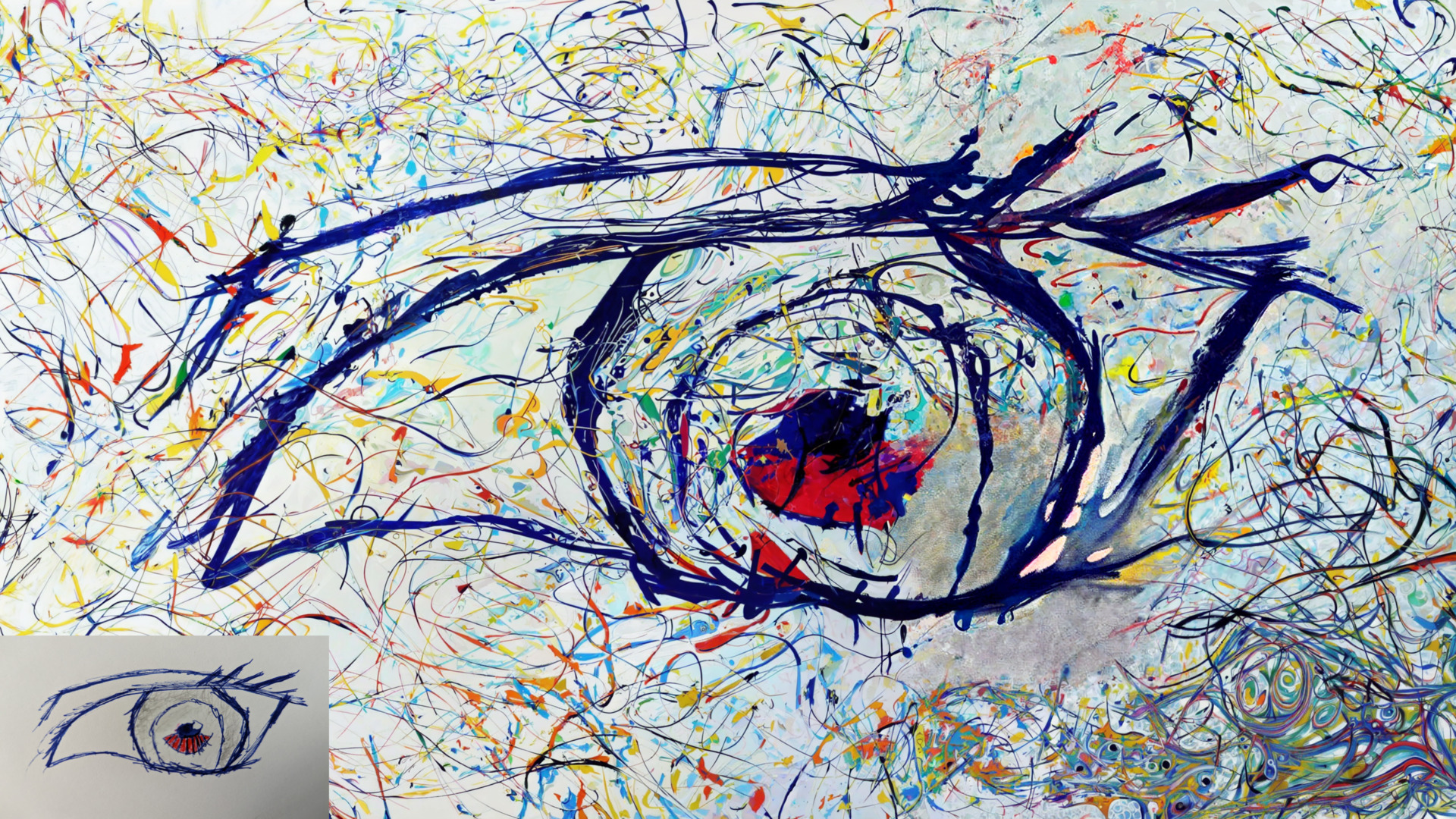}
    \caption{4K image generated with 8x11 models arranged in a grid, using a sketch (shown in bottom-left corner) as a guide image with $25\%$ strength for the whole canvas. All the models use the prompt 
    "Abstract decorative illustration, by jackson pollock, elegant, intricate, highly detailed, smooth, sharp focus, vibrant colors, artstation, stunning masterpiece". Full resolution image available at \url{https://albarji-mixture-of-diffusers-paper.s3.eu-west-1.amazonaws.com/eyeguided.png}.}
    \label{fig:4Kguided}
\end{figure}

Using a guide image can also be exploited to further enforce coherence across many diffusion regions. As an example, Figure \ref{fig:4Kguided} presents a 4K image for which a sketch has been used as guidance image for the full canvas. The result preserves the overall structure, while adding patterns that follow the text prompts of the diffusion models.

\section{Discussion and further work}

In this paper we have presented Mixture of Diffusers, a new strategy for combining several diffusion models that act on the same canvas to generate a single image. The method provides several advantages over using a single diffusion model, such as a more detailed control on image composition, the ability to generate high resolution images within a small GPU RAM budget, and more flexibility for "image2image" or image-guided generation.

A current limitation of this method is that each diffusion model is restricted to act on a rectangular region. Although this restriction is inherited from the workings of the actual noise prediction models, trained over rectangular images, it might be possible to use some masking similar to existing inpainting techniques to influence free-from regions, thus providing further control on the location of the objects described in the prompt. This and other possible extensions could be the aim of further work on this method.

\section*{Acknowledgements}

We would like to thank the Instituto de Ingeniería del Conocimiento for providing the hardware resources for the experiments in this paper, as well as for the discussions with many colleagues. Hardware resources were also partially funded by projects PID2019-106827GB-I00 / AEI / 10.13039/501100011033.

\bibliographystyle{unsrt}  
\bibliography{references}

\appendix

\newpage
\section{Additional image composition results}
\label{app:composition}

\begin{figure}
    \centering
    \begin{subfigure}[b]{0.33\textwidth}
        \centering
        \includegraphics[width=\textwidth]{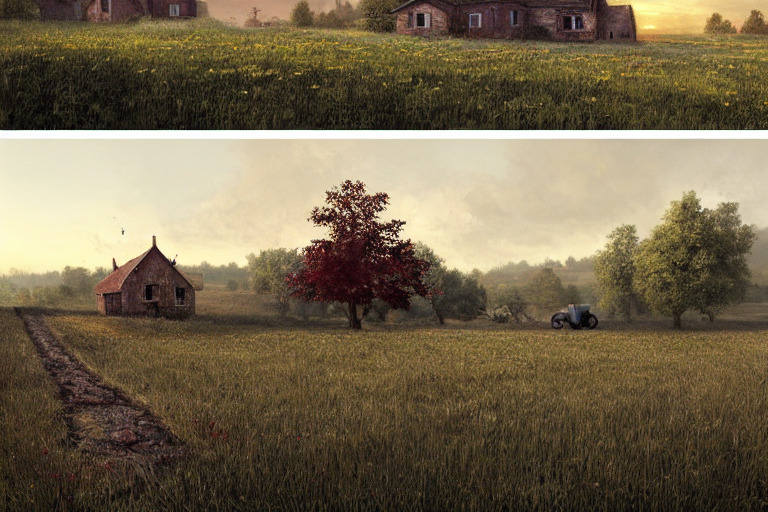}
    \end{subfigure}
    \begin{subfigure}[b]{0.33\textwidth}
        \centering
        \includegraphics[width=\textwidth]{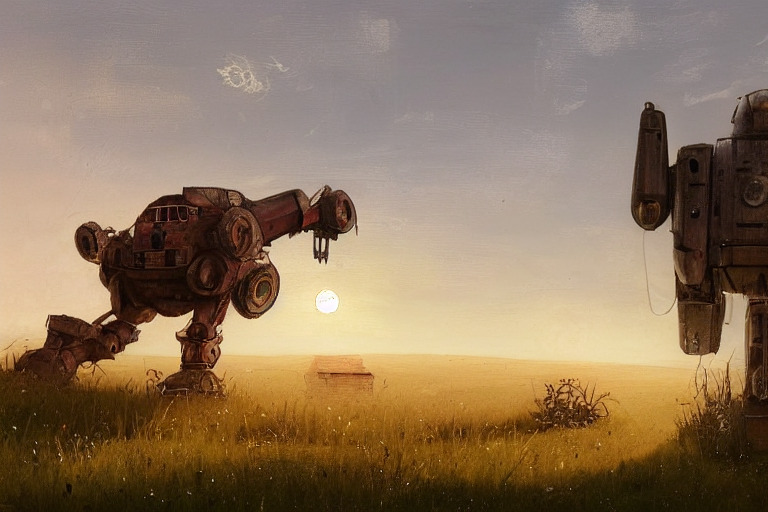}
    \end{subfigure}
    \begin{subfigure}[b]{0.33\textwidth}
        \centering
        \includegraphics[width=\textwidth]{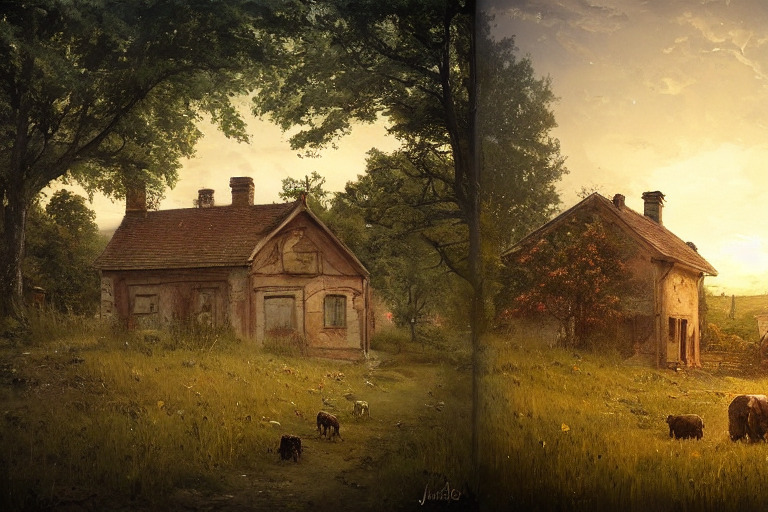}
    \end{subfigure}
    \begin{subfigure}[b]{0.33\textwidth}
        \centering
        \includegraphics[width=\textwidth]{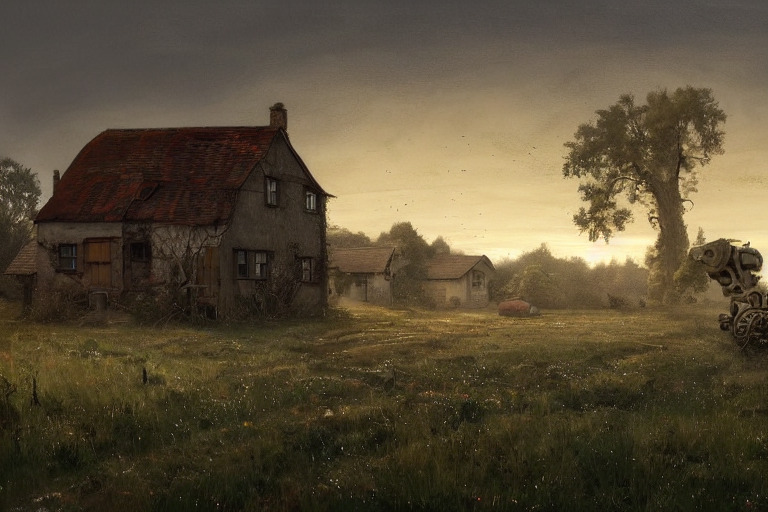}
    \end{subfigure}
    \begin{subfigure}[b]{0.33\textwidth}
        \centering
        \includegraphics[width=\textwidth]{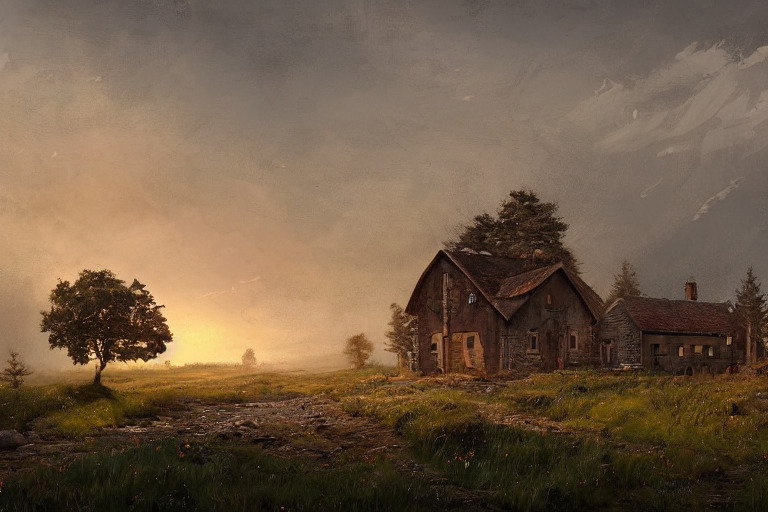}
    \end{subfigure}
    \begin{subfigure}[b]{0.33\textwidth}
        \centering
        \includegraphics[width=\textwidth]{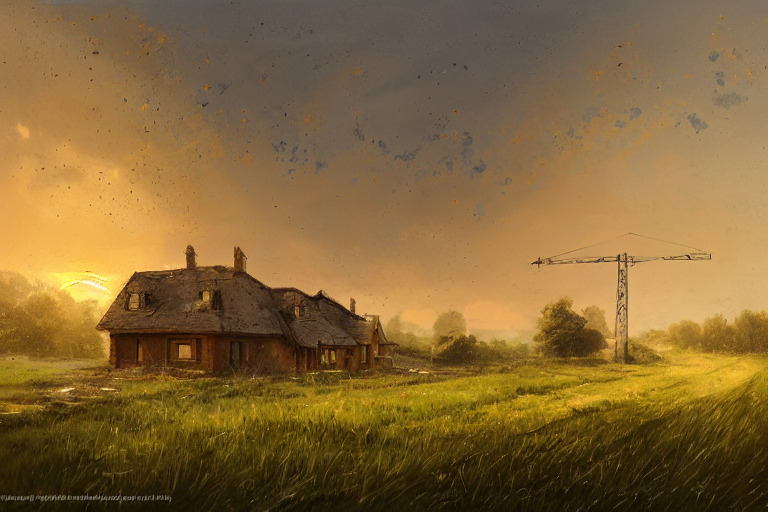}
    \end{subfigure}
    \begin{subfigure}[b]{0.33\textwidth}
        \centering
        \includegraphics[width=\textwidth]{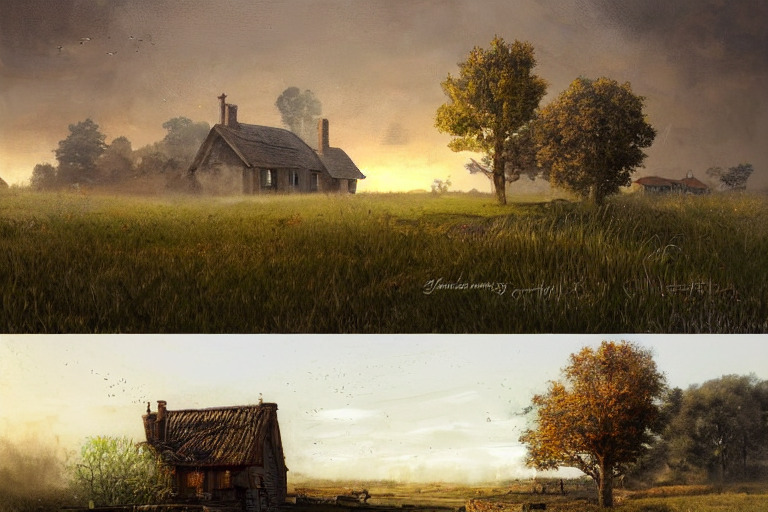}
    \end{subfigure}
    \begin{subfigure}[b]{0.33\textwidth}
        \centering
        \includegraphics[width=\textwidth]{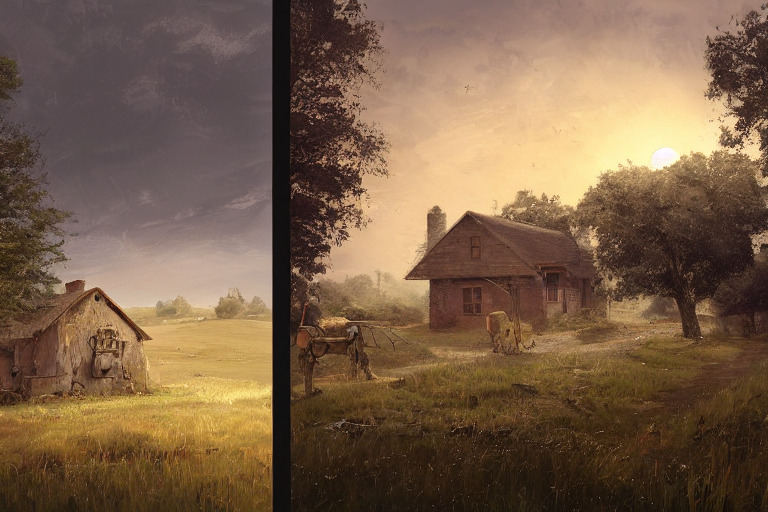}
    \end{subfigure}
    \begin{subfigure}[b]{0.33\textwidth}
        \centering
        \includegraphics[width=\textwidth]{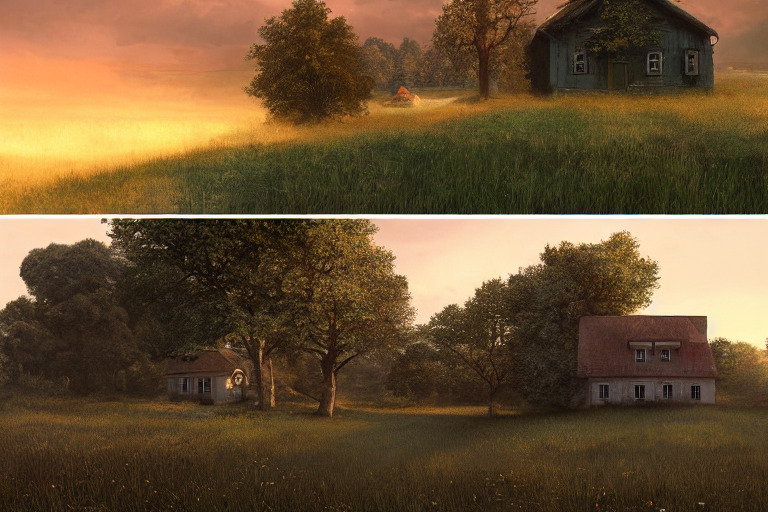}
    \end{subfigure}
    \begin{subfigure}[b]{0.33\textwidth}
        \centering
        \includegraphics[width=\textwidth]{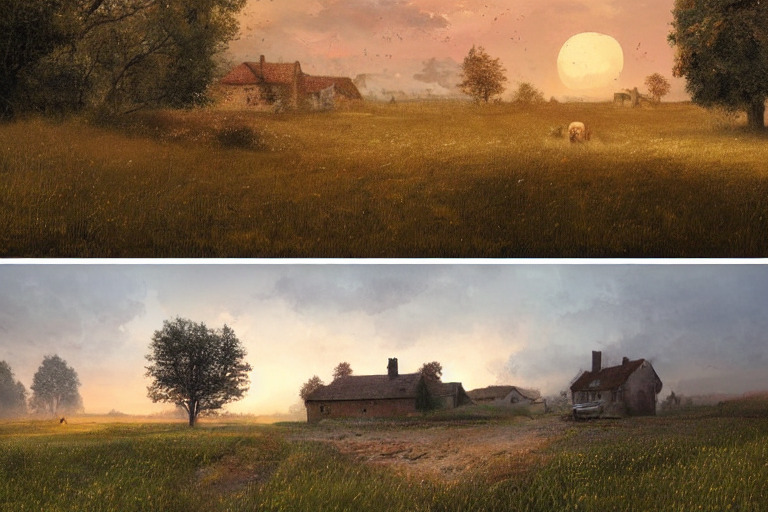}
    \end{subfigure}
    \begin{subfigure}[b]{0.33\textwidth}
        \centering
        \includegraphics[width=\textwidth]{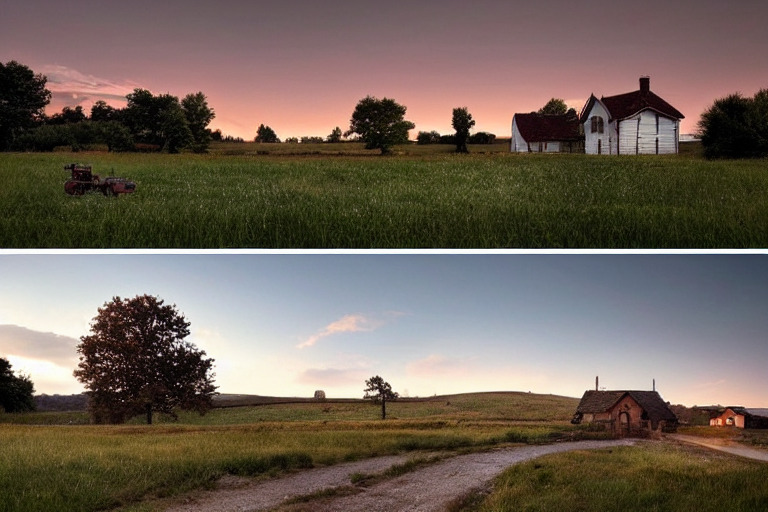}
    \end{subfigure}
    \begin{subfigure}[b]{0.33\textwidth}
        \centering
        \includegraphics[width=\textwidth]{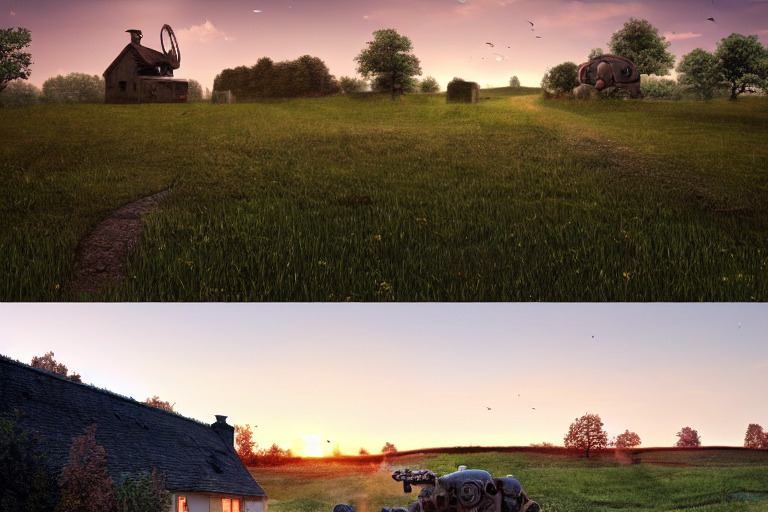}
    \end{subfigure}
    \begin{subfigure}[b]{0.33\textwidth}
        \centering
        \includegraphics[width=\textwidth]{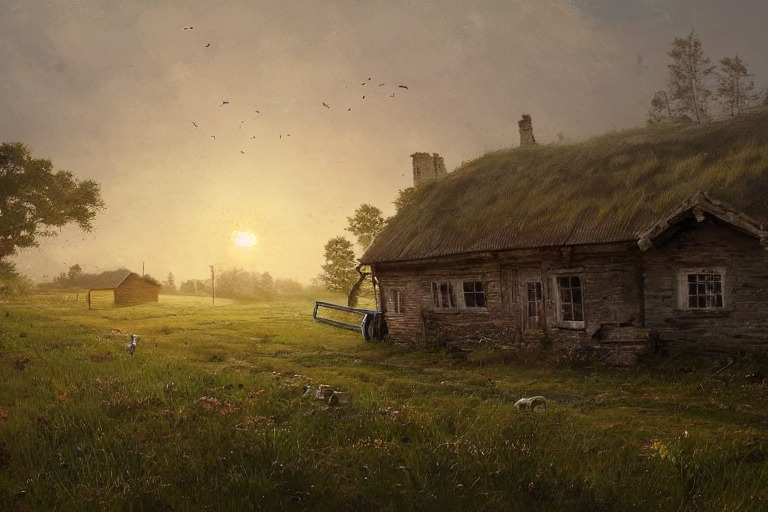}
    \end{subfigure}
    \begin{subfigure}[b]{0.33\textwidth}
        \centering
        \includegraphics[width=\textwidth]{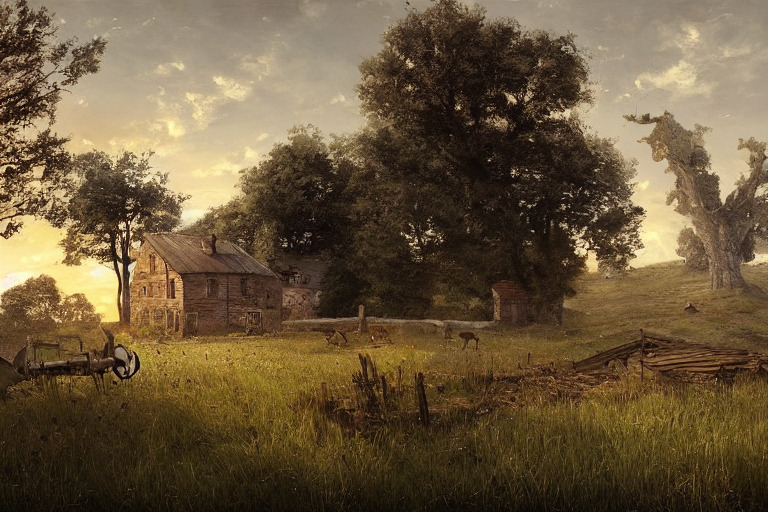}
    \end{subfigure}
    \begin{subfigure}[b]{0.33\textwidth}
        \centering
        \includegraphics[width=\textwidth]{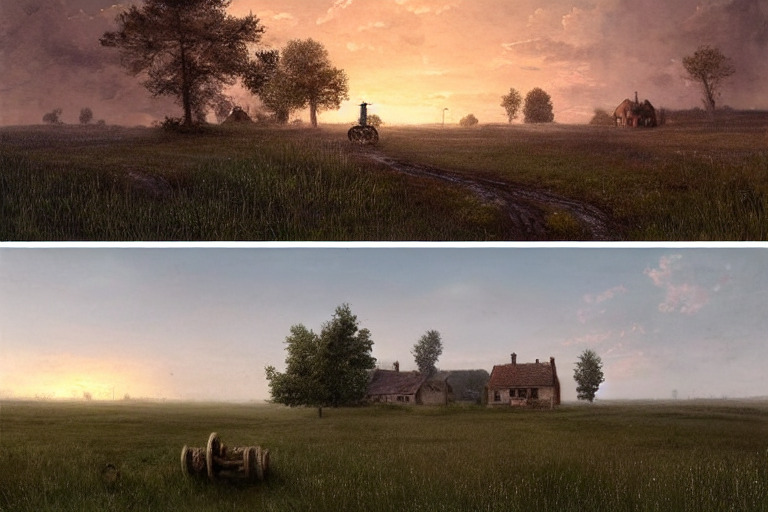}
    \end{subfigure}
    \begin{subfigure}[b]{0.33\textwidth}
        \centering
        \includegraphics[width=\textwidth]{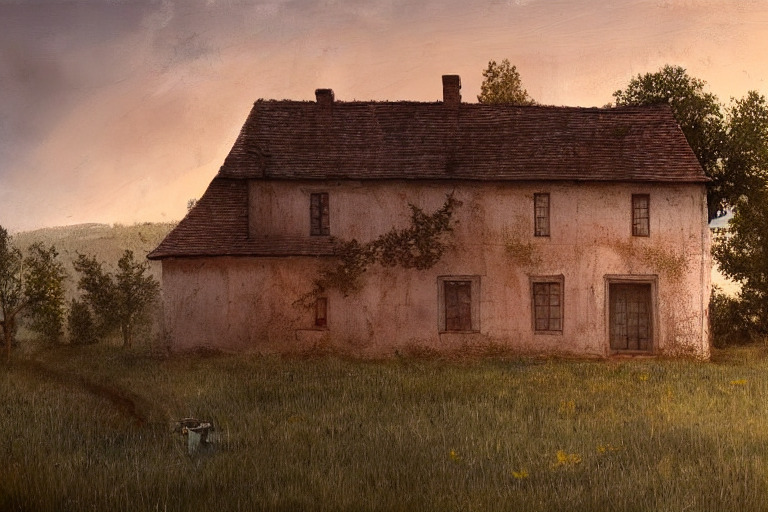}
    \end{subfigure}
    \begin{subfigure}[b]{0.33\textwidth}
        \centering
        \includegraphics[width=\textwidth]{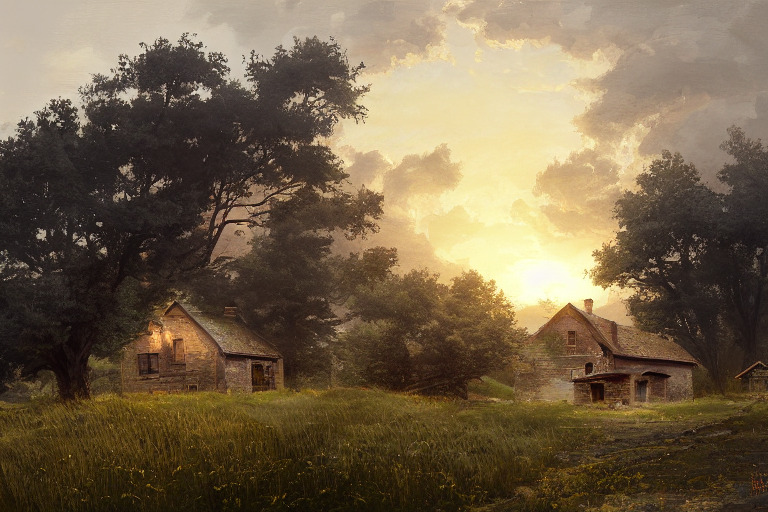}
    \end{subfigure}
    \begin{subfigure}[b]{0.33\textwidth}
        \centering
        \includegraphics[width=\textwidth]{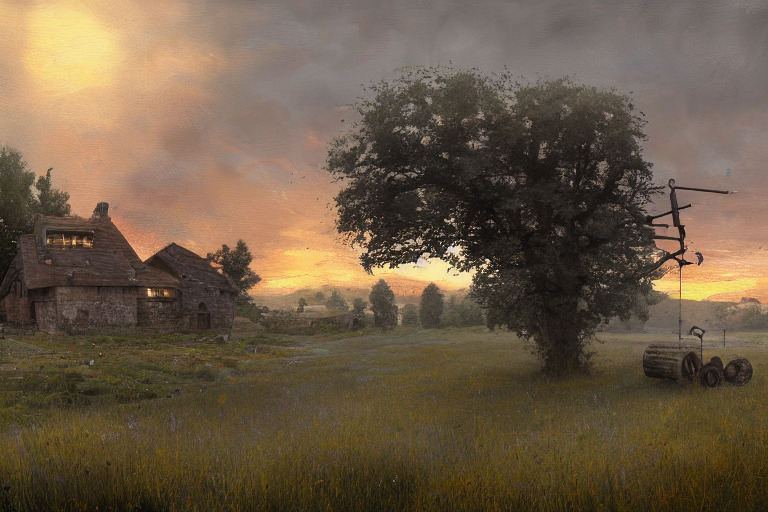}
    \end{subfigure}
    \caption{Unfiltered samples from Stable Diffusion 1.4 for the \textbf{Unique} prompt in Figure \ref{fig:compositionPrompts}.}
    \label{fig:compositionsd}
\end{figure}

\begin{figure}
    \centering
    \begin{subfigure}[b]{0.33\textwidth}
        \centering
        \includegraphics[width=\textwidth]{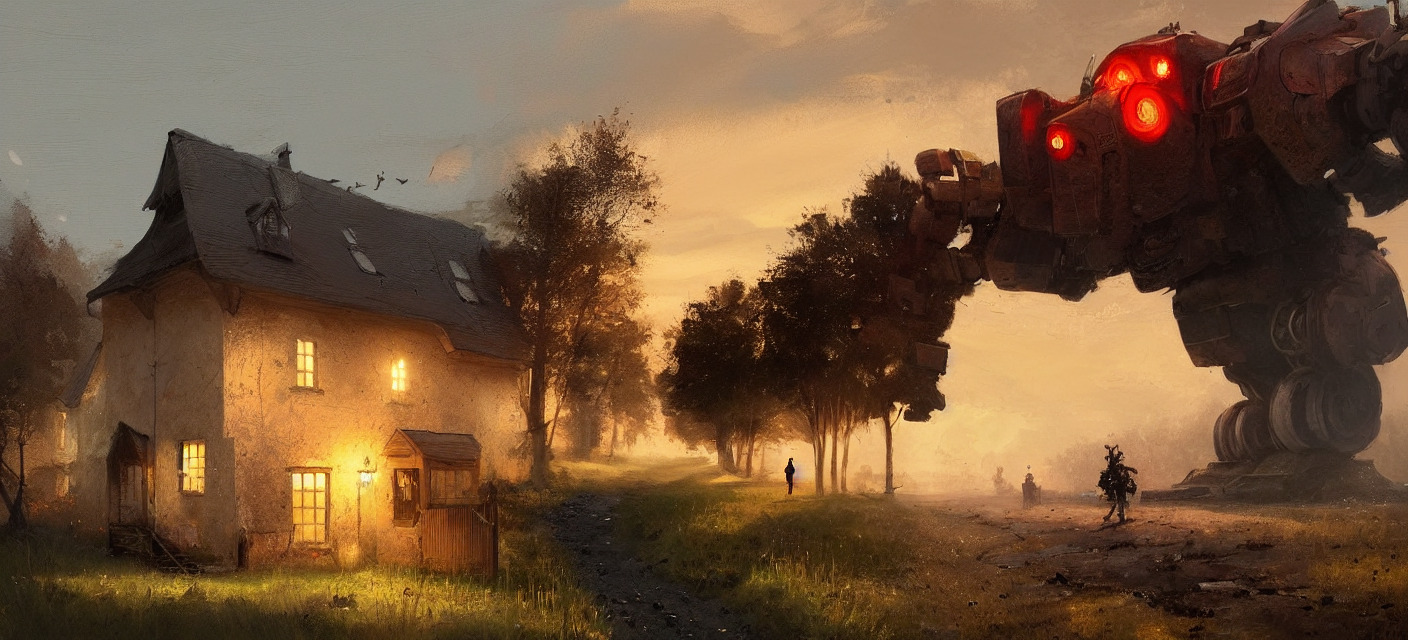}
    \end{subfigure}
    \begin{subfigure}[b]{0.33\textwidth}
        \centering
        \includegraphics[width=\textwidth]{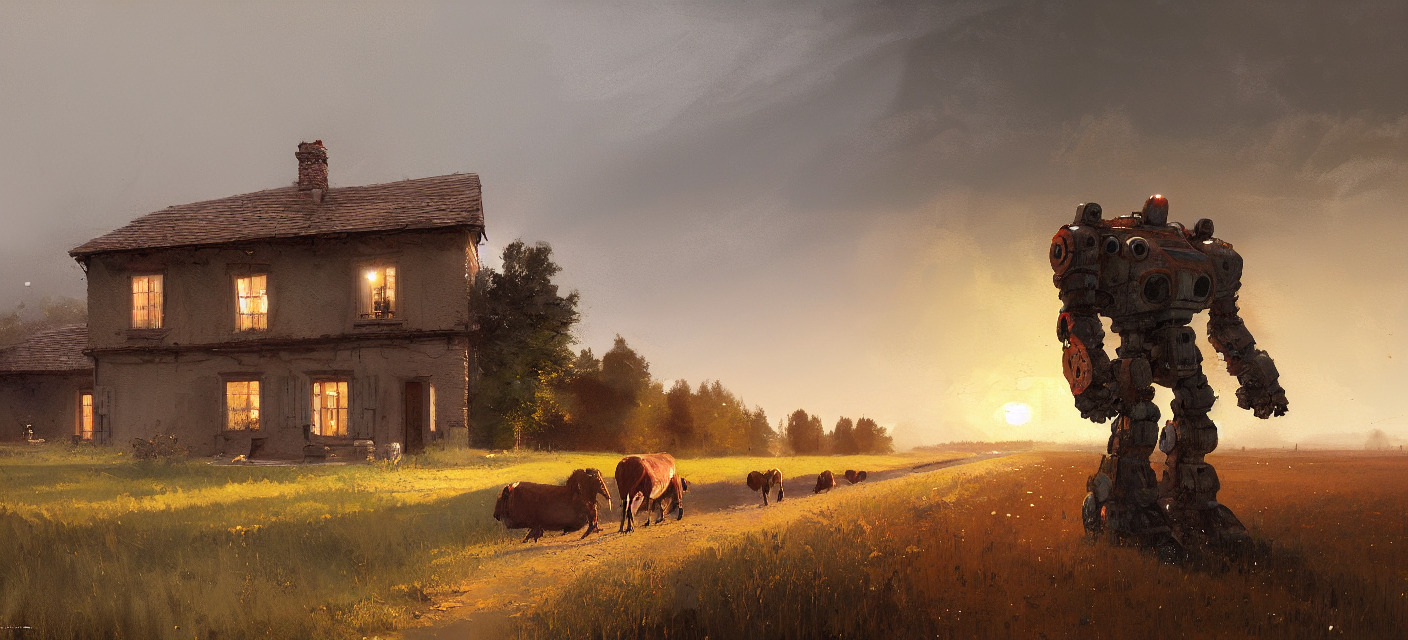}
    \end{subfigure}
    \begin{subfigure}[b]{0.33\textwidth}
        \centering
        \includegraphics[width=\textwidth]{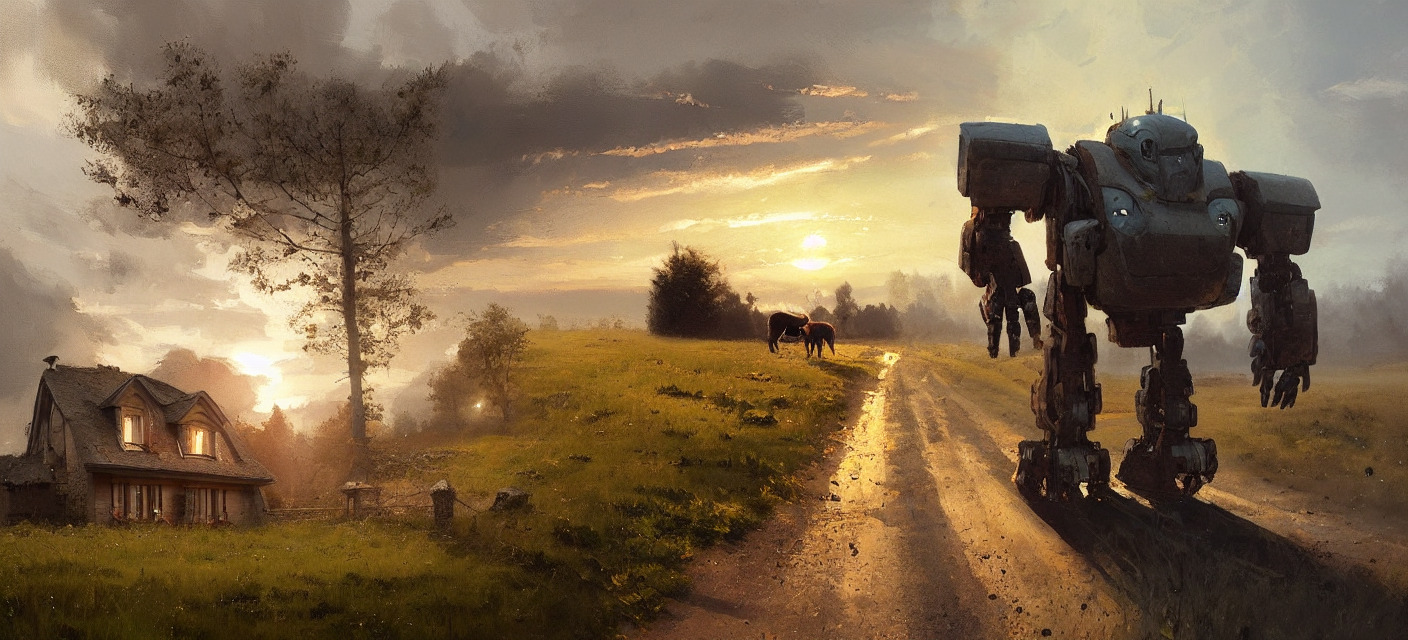}
    \end{subfigure}
    \begin{subfigure}[b]{0.33\textwidth}
        \centering
        \includegraphics[width=\textwidth]{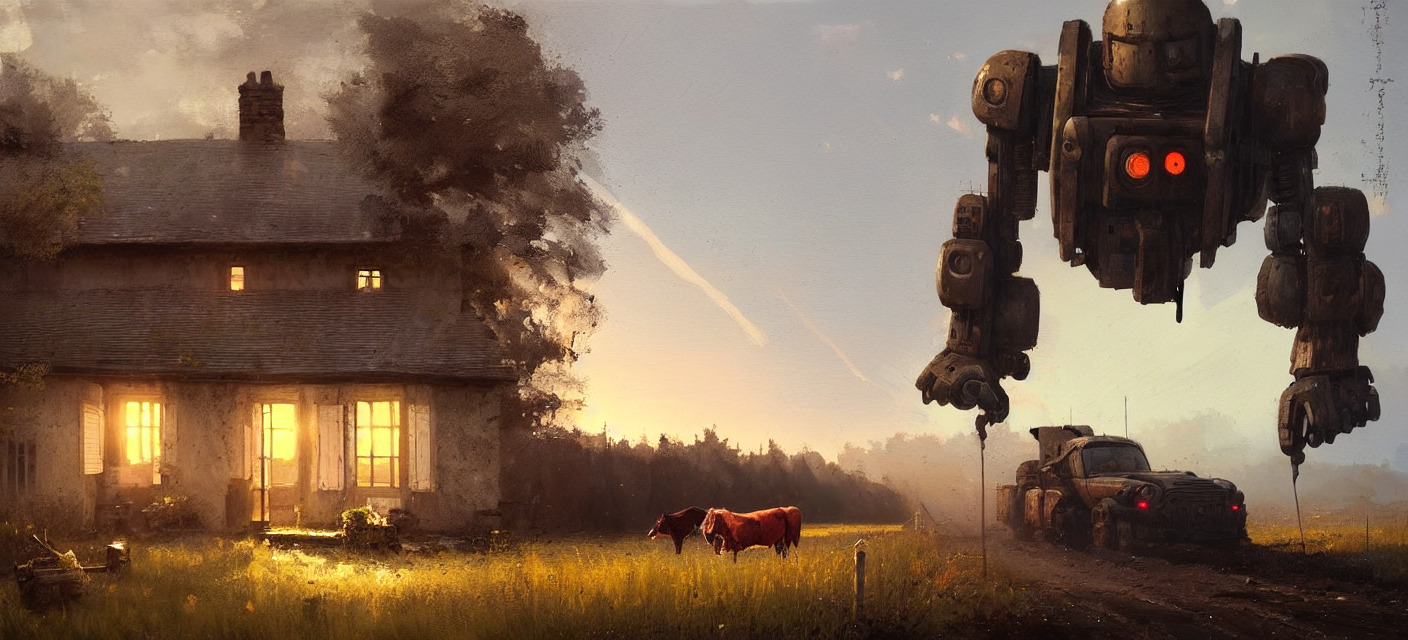}
    \end{subfigure}
    \begin{subfigure}[b]{0.33\textwidth}
        \centering
        \includegraphics[width=\textwidth]{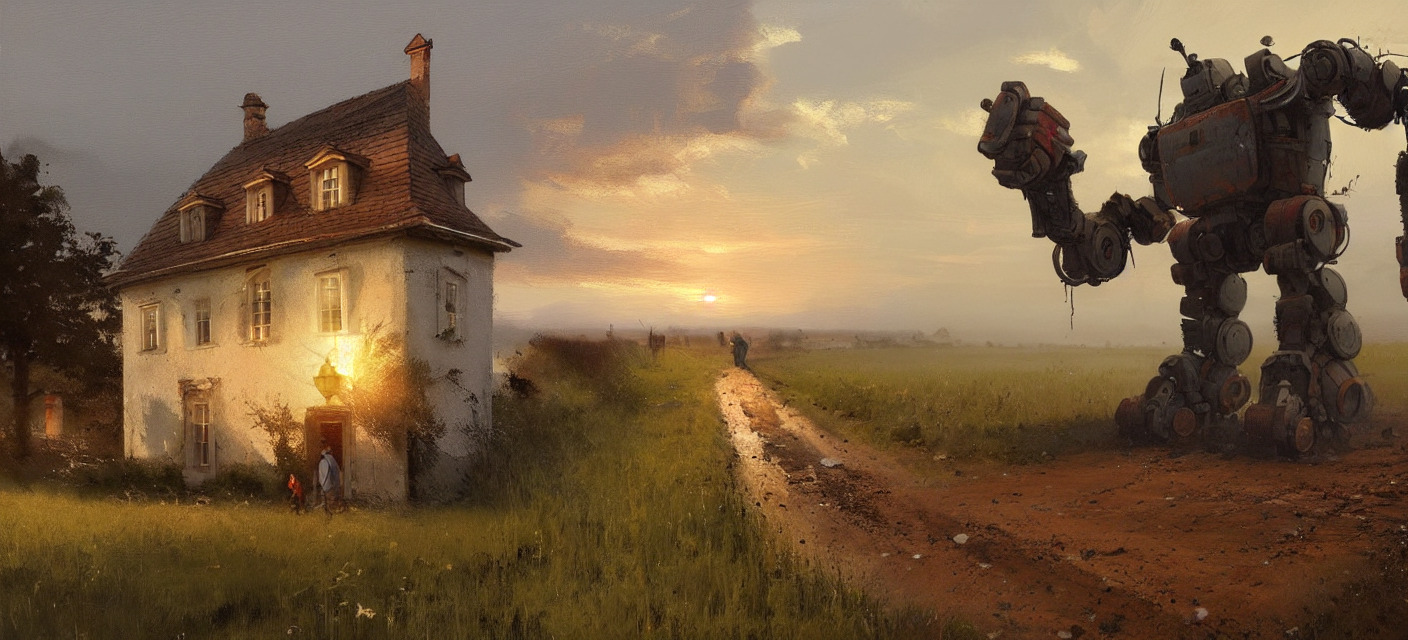}
    \end{subfigure}
    \begin{subfigure}[b]{0.33\textwidth}
        \centering
        \includegraphics[width=\textwidth]{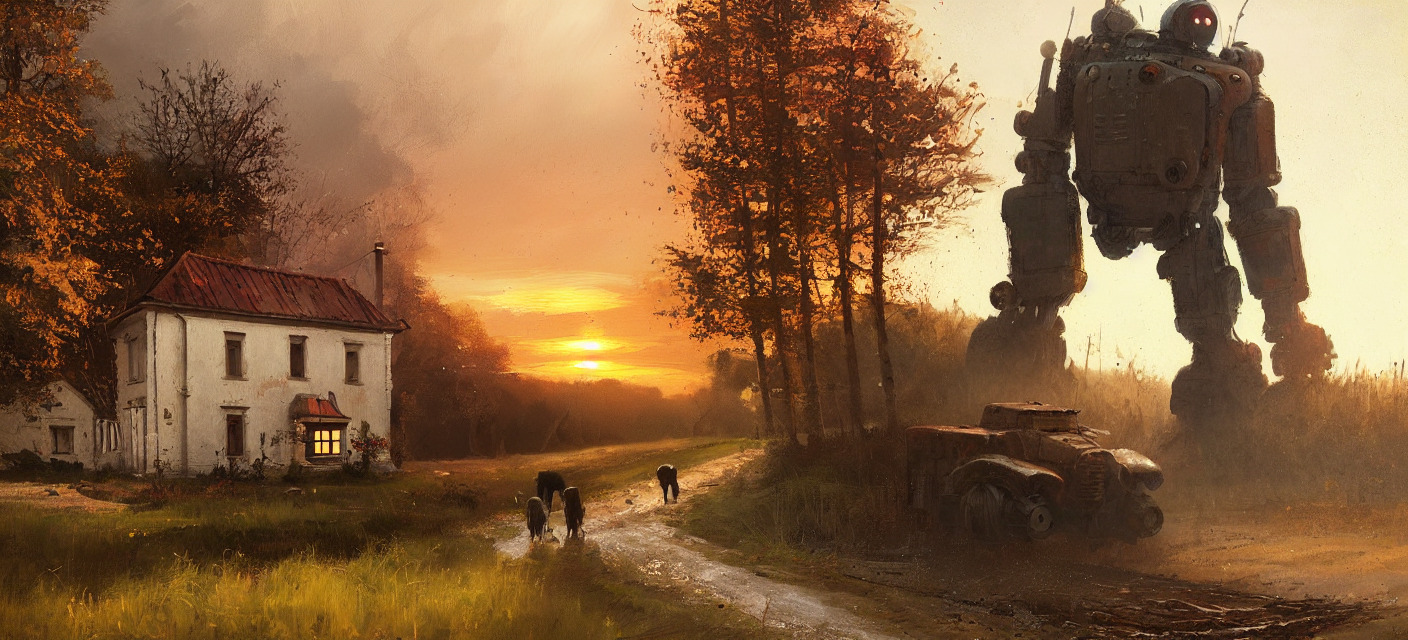}
    \end{subfigure}
    \begin{subfigure}[b]{0.33\textwidth}
        \centering
        \includegraphics[width=\textwidth]{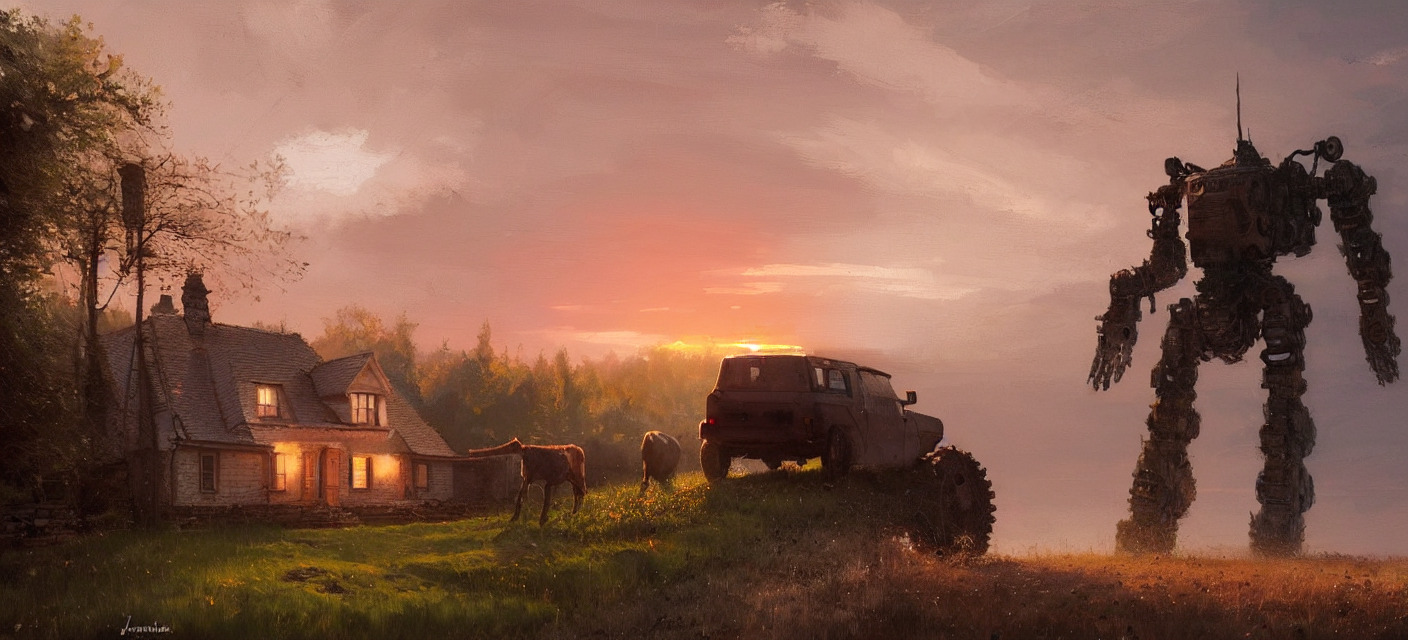}
    \end{subfigure}
    \begin{subfigure}[b]{0.33\textwidth}
        \centering
        \includegraphics[width=\textwidth]{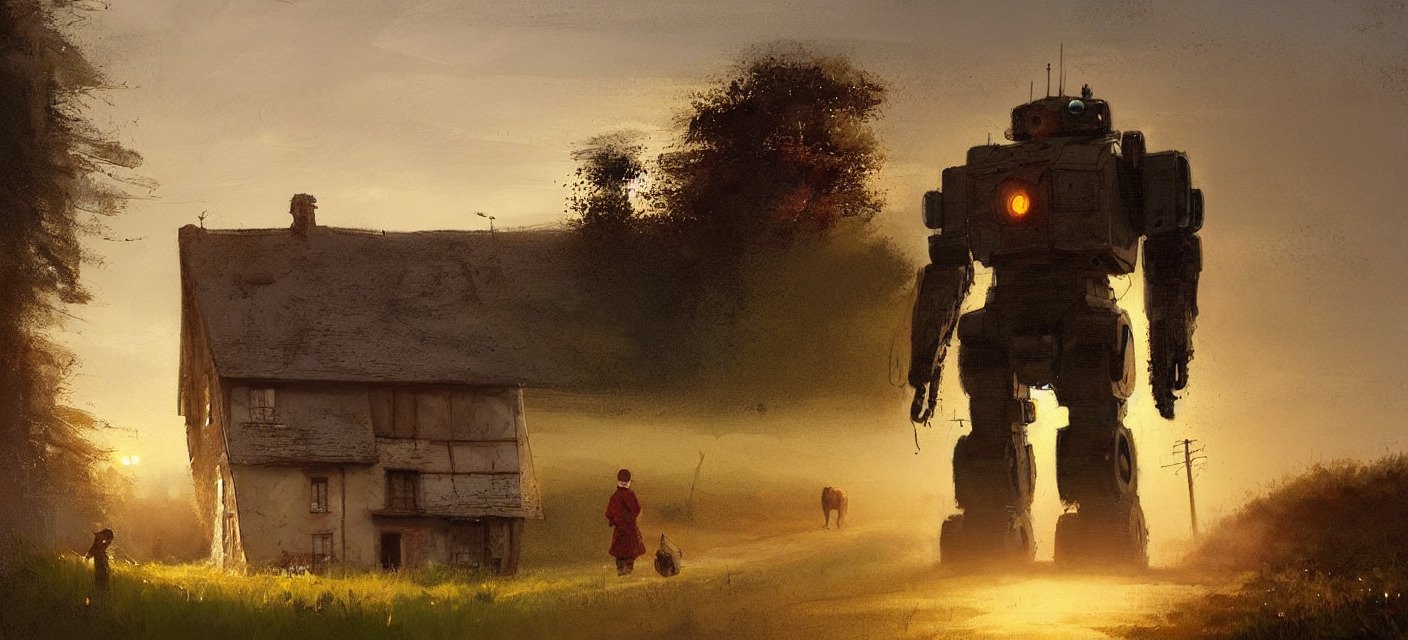}
    \end{subfigure}
    \begin{subfigure}[b]{0.33\textwidth}
        \centering
        \includegraphics[width=\textwidth]{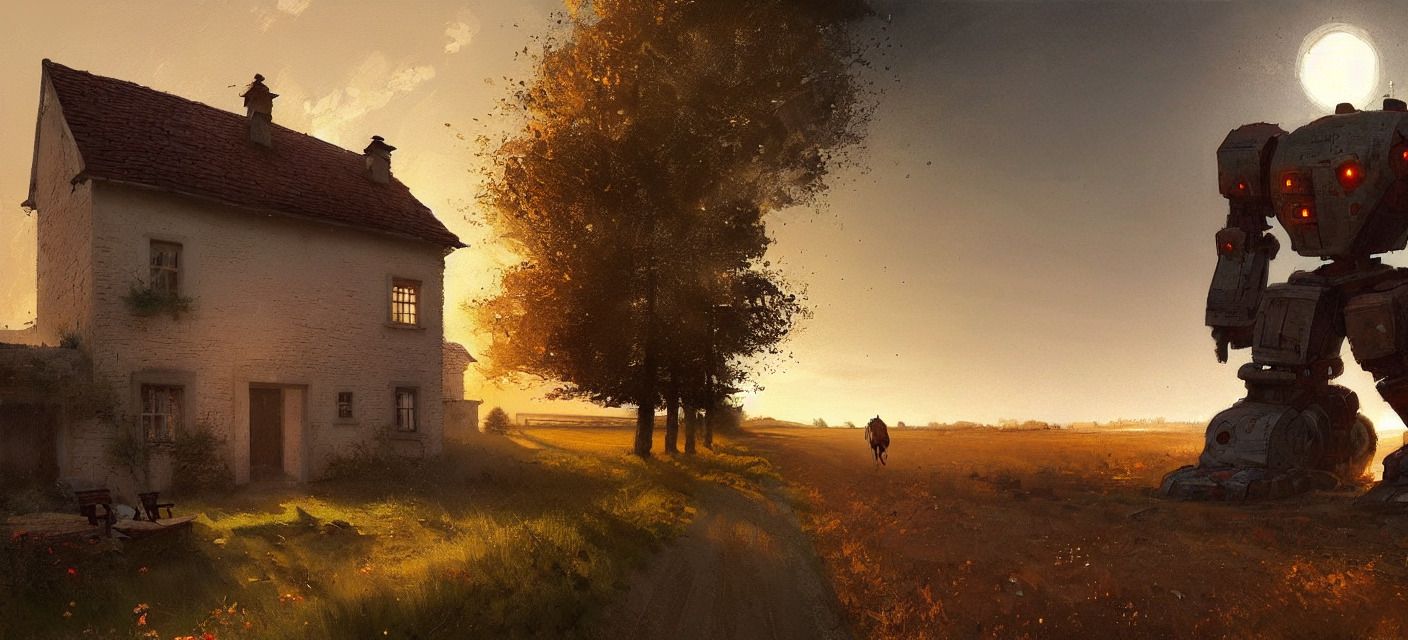}
    \end{subfigure}
    \begin{subfigure}[b]{0.33\textwidth}
        \centering
        \includegraphics[width=\textwidth]{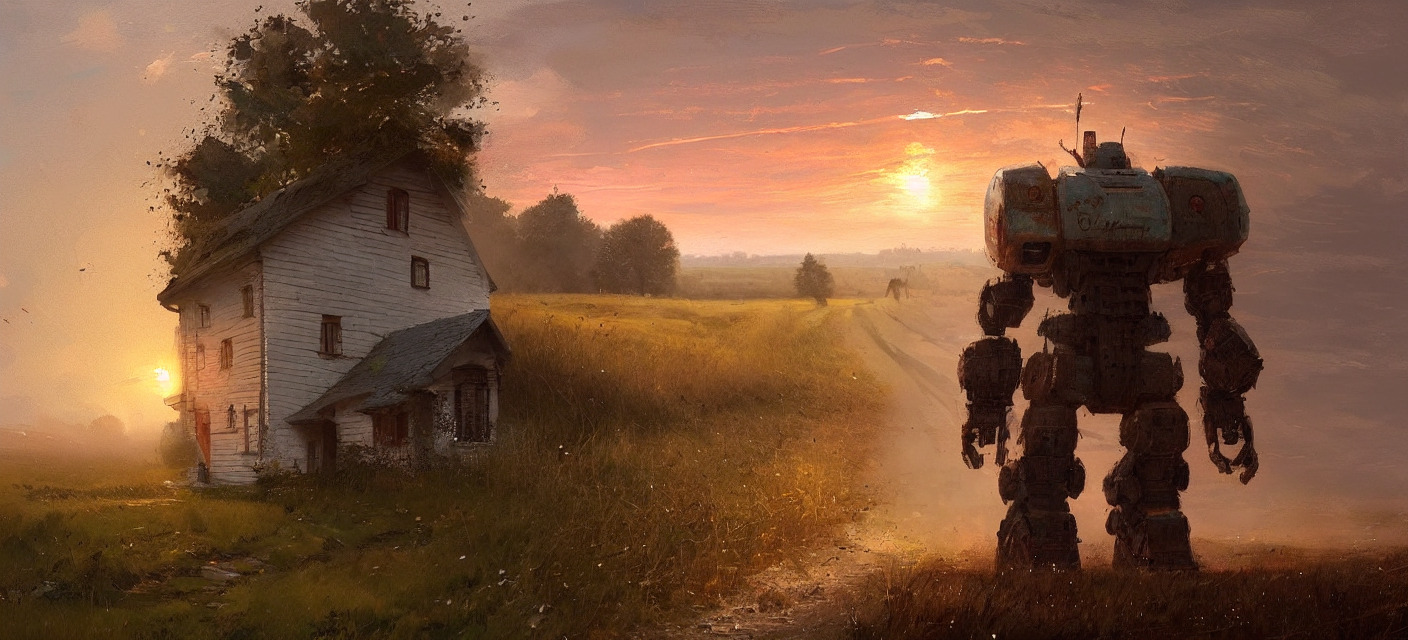}
    \end{subfigure}
    \begin{subfigure}[b]{0.33\textwidth}
        \centering
        \includegraphics[width=\textwidth]{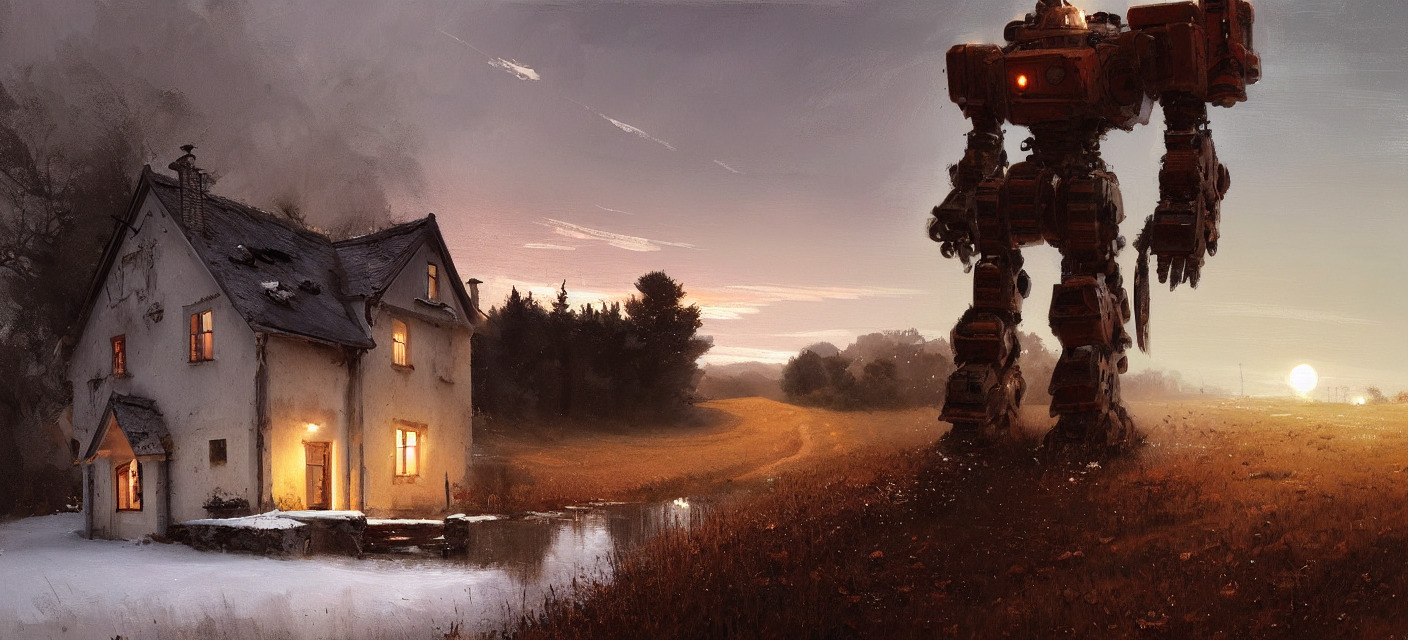}
    \end{subfigure}
    \begin{subfigure}[b]{0.33\textwidth}
        \centering
        \includegraphics[width=\textwidth]{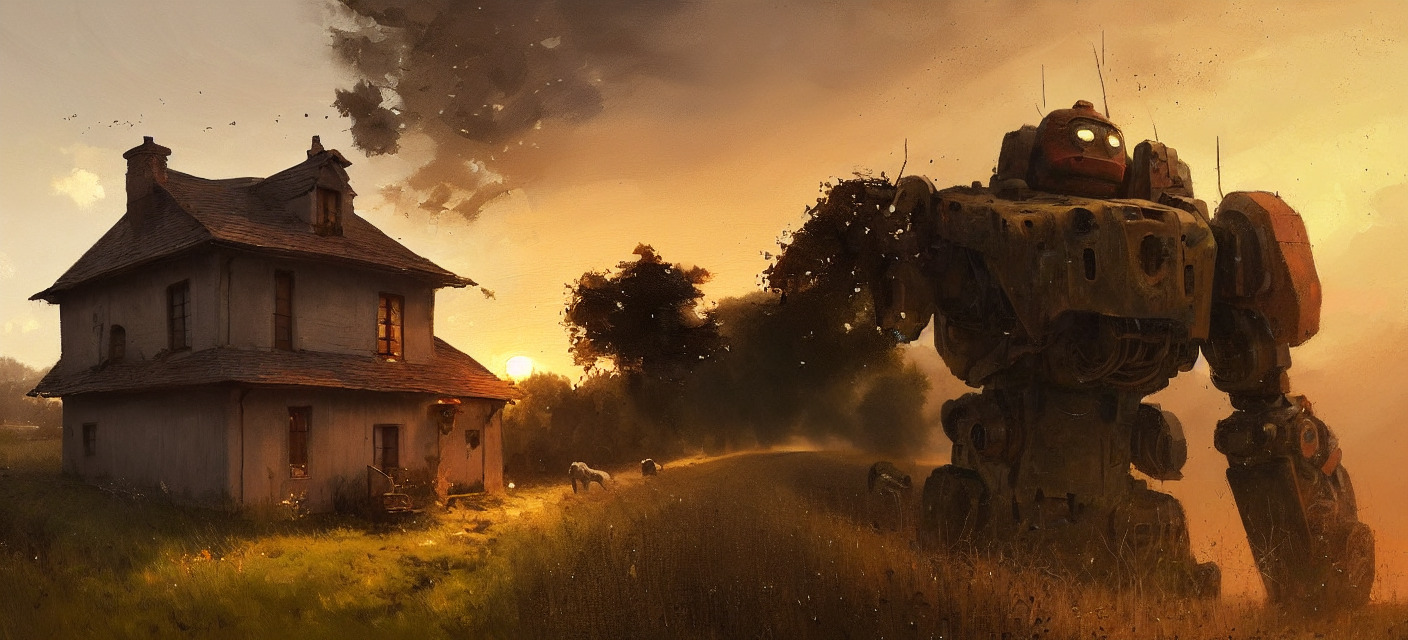}
    \end{subfigure}
    \begin{subfigure}[b]{0.33\textwidth}
        \centering
        \includegraphics[width=\textwidth]{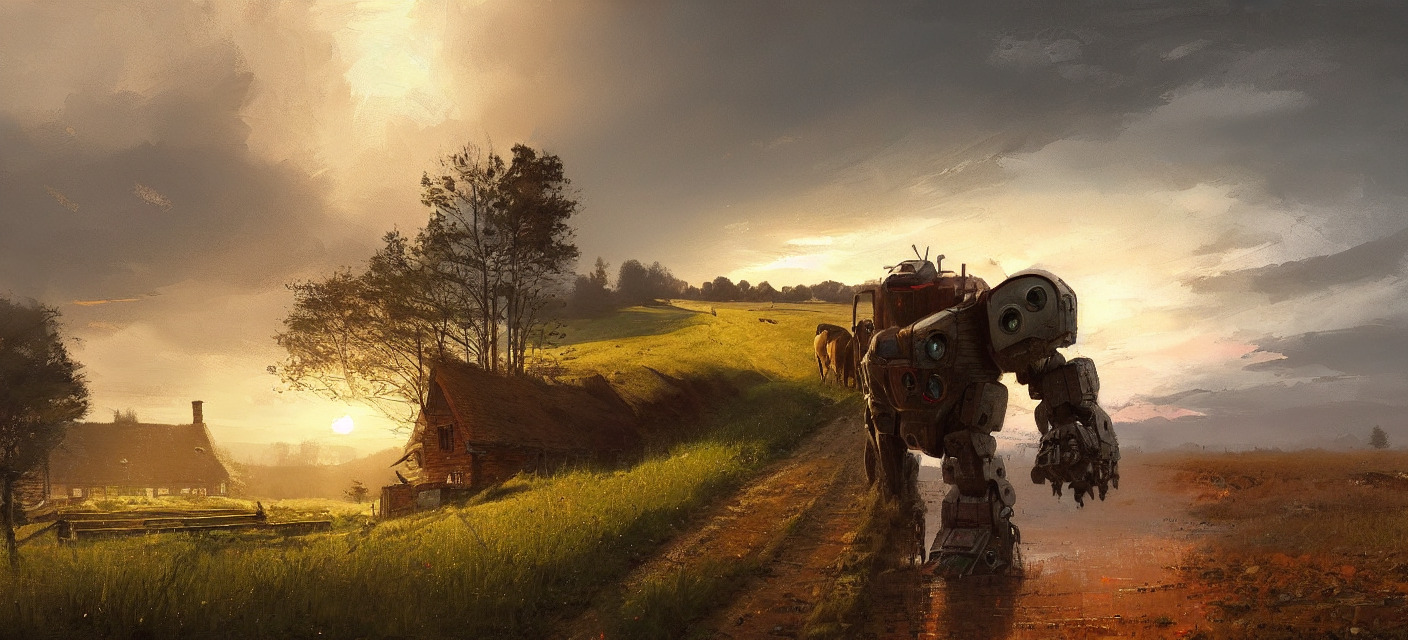}
    \end{subfigure}
    \begin{subfigure}[b]{0.33\textwidth}
        \centering
        \includegraphics[width=\textwidth]{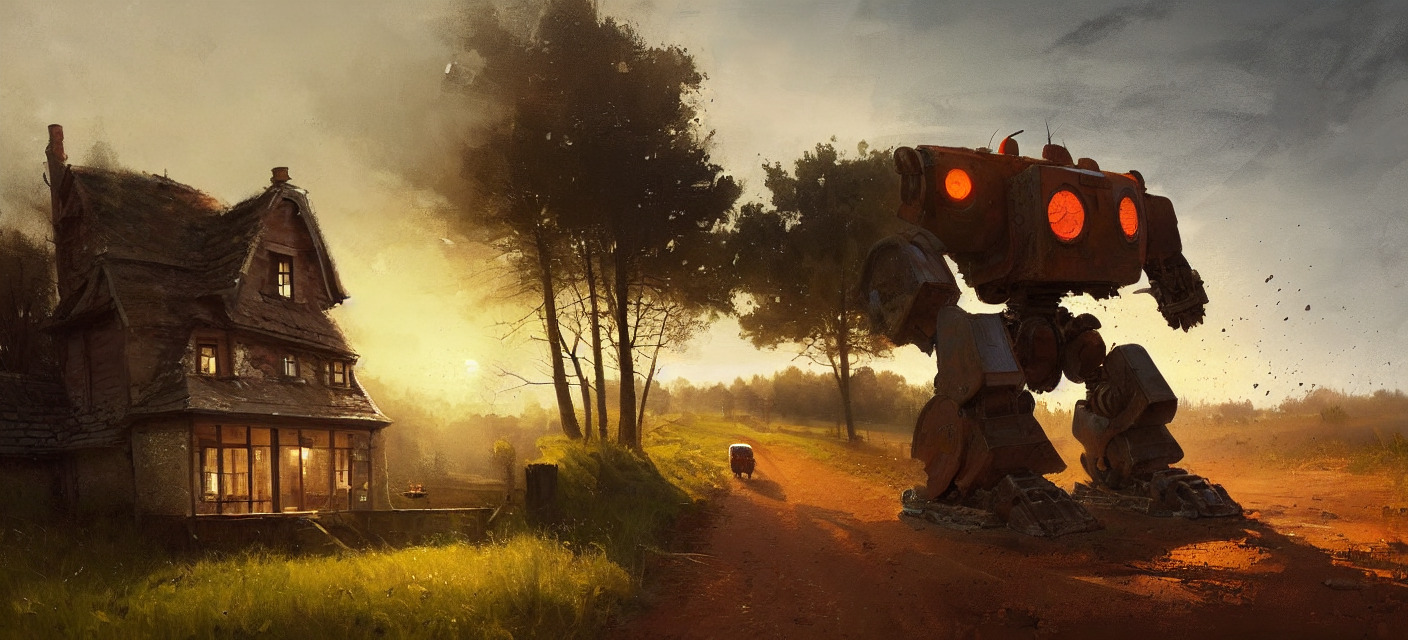}
    \end{subfigure}
    \begin{subfigure}[b]{0.33\textwidth}
        \centering
        \includegraphics[width=\textwidth]{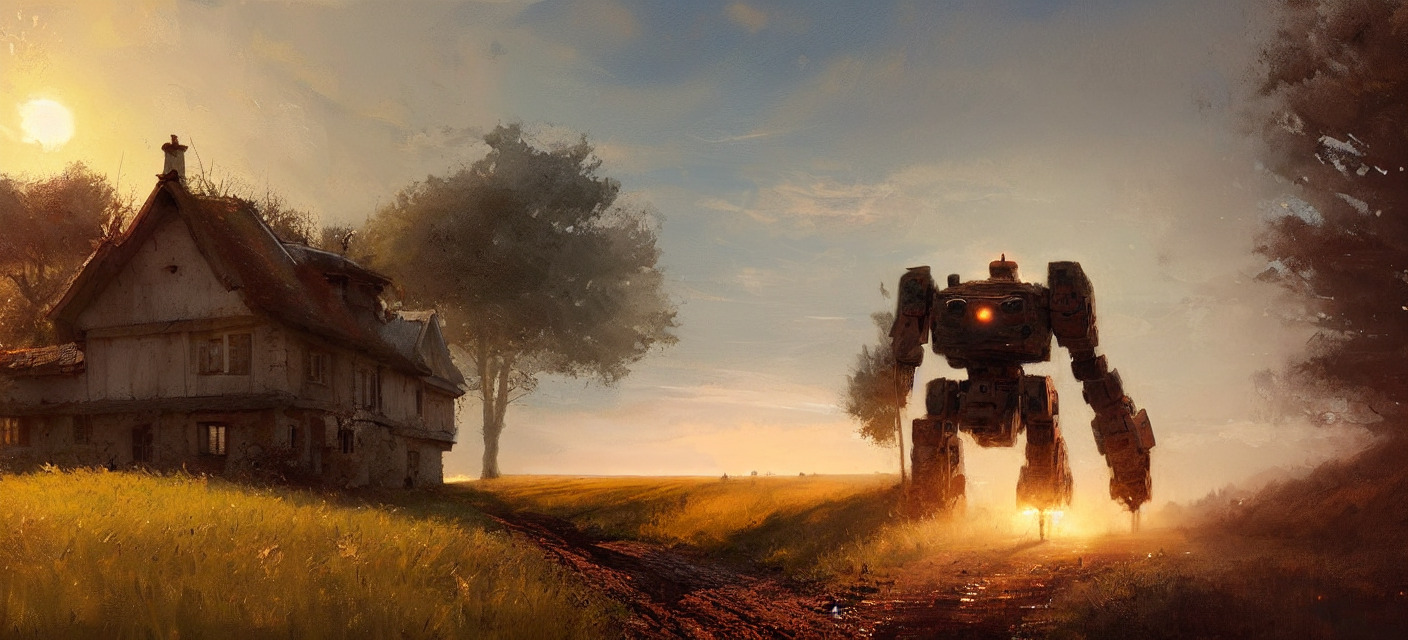}
    \end{subfigure}
    \begin{subfigure}[b]{0.33\textwidth}
        \centering
        \includegraphics[width=\textwidth]{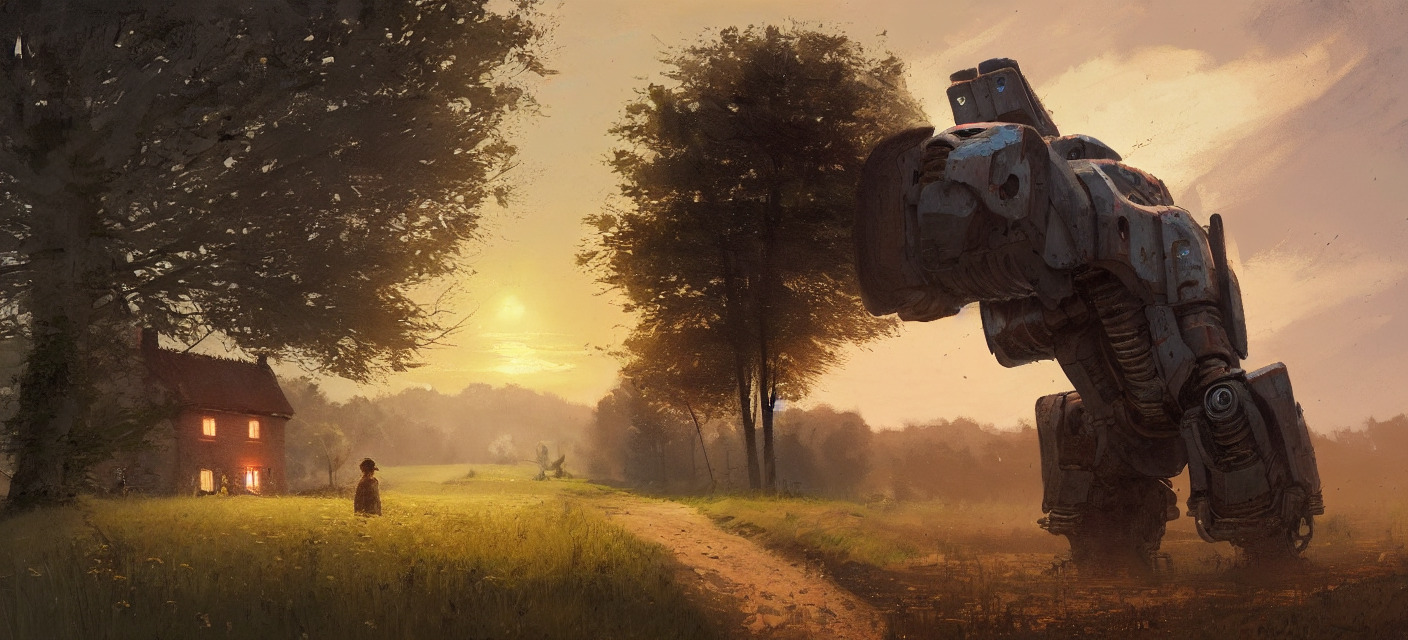}
    \end{subfigure}
    \begin{subfigure}[b]{0.33\textwidth}
        \centering
        \includegraphics[width=\textwidth]{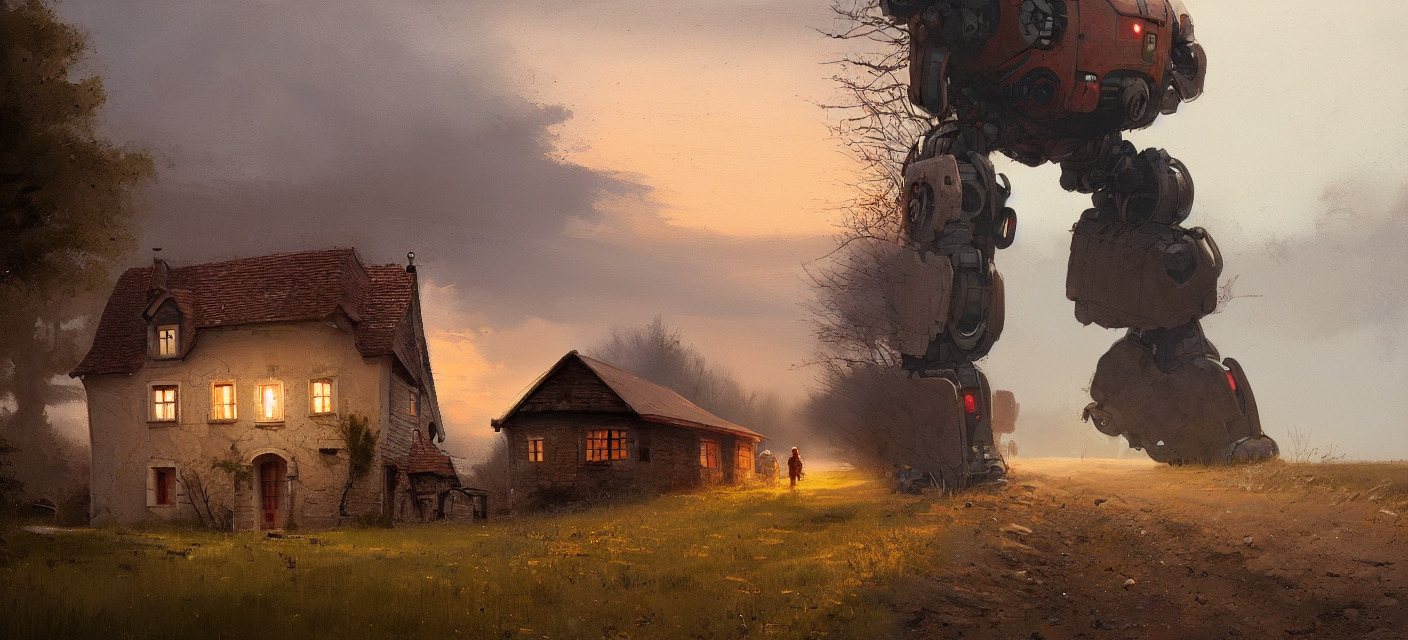}
    \end{subfigure}
    \begin{subfigure}[b]{0.33\textwidth}
        \centering
        \includegraphics[width=\textwidth]{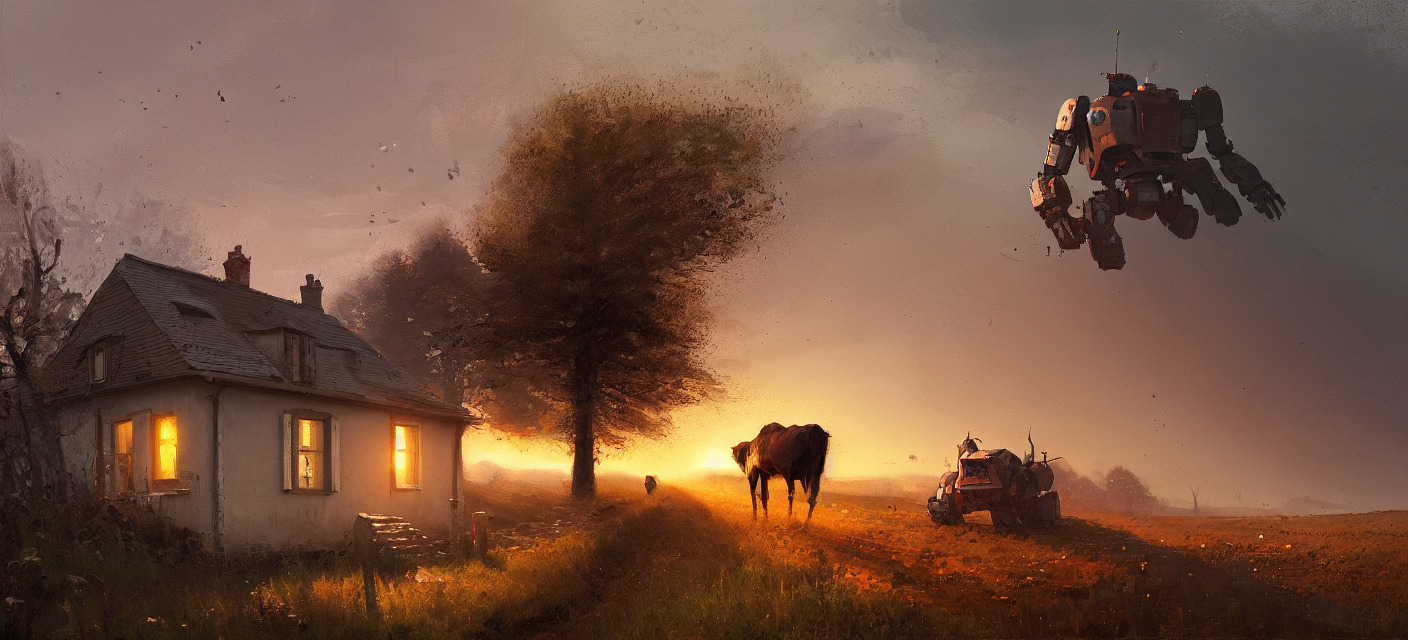}
    \end{subfigure}
    \caption{Unfiltered samples from Mixture of Diffusers using the \textbf{Left}, \textbf{Center} and \textbf{Right} prompts from Figure \ref{fig:compositionPrompts}.}
    \label{fig:compositionmod}
\end{figure}

Figures \ref{fig:compositionsd} and \ref{fig:compositionmod} present random samples (e.g. not selected or filtered in any way) from Stable Diffusion 1.4 and Mixture of Diffusers, showing how the presented method is able to better represent the intended composition of the user.

\section{Additional 4K results}
\label{app:4k}

\begin{figure}
    \centering
    \includegraphics[width=\textwidth]{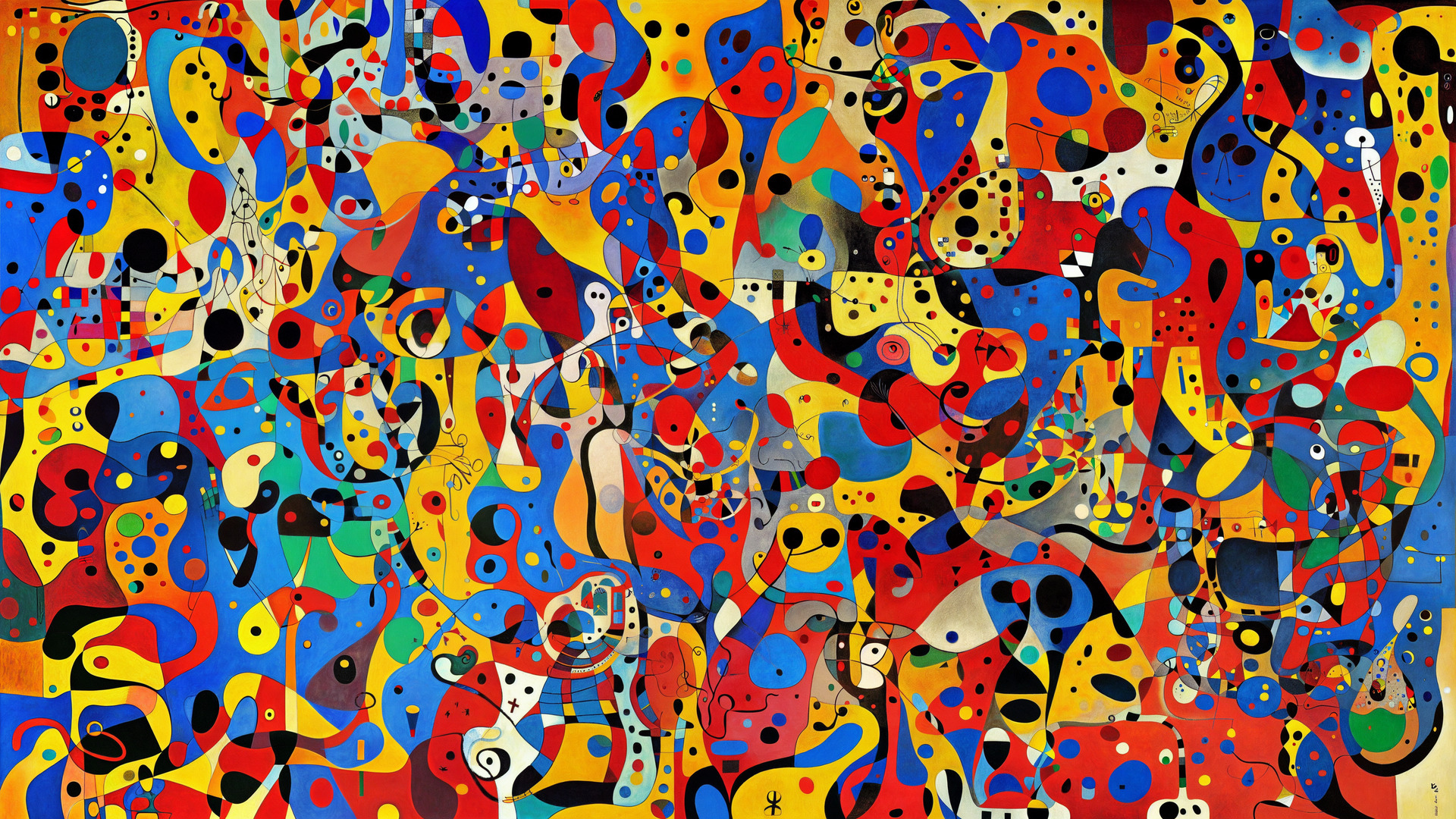}
    \caption{4K image generated with 8x11 models arranged in a grid, all of them using the prompt "Abstract decorative illustration, by joan miro and gustav klimt and marlina vera and loish, elegant, intricate, highly detailed, smooth, sharp focus, vibrant colors, artstation, stunning masterpiece". Full resolution image available at \url{https://albarji-mixture-of-diffusers-paper.s3.eu-west-1.amazonaws.com/4Kmiro.png}.}
    \label{fig:4kmiro}
\end{figure}

\begin{figure}
    \centering
    \includegraphics[width=\textwidth]{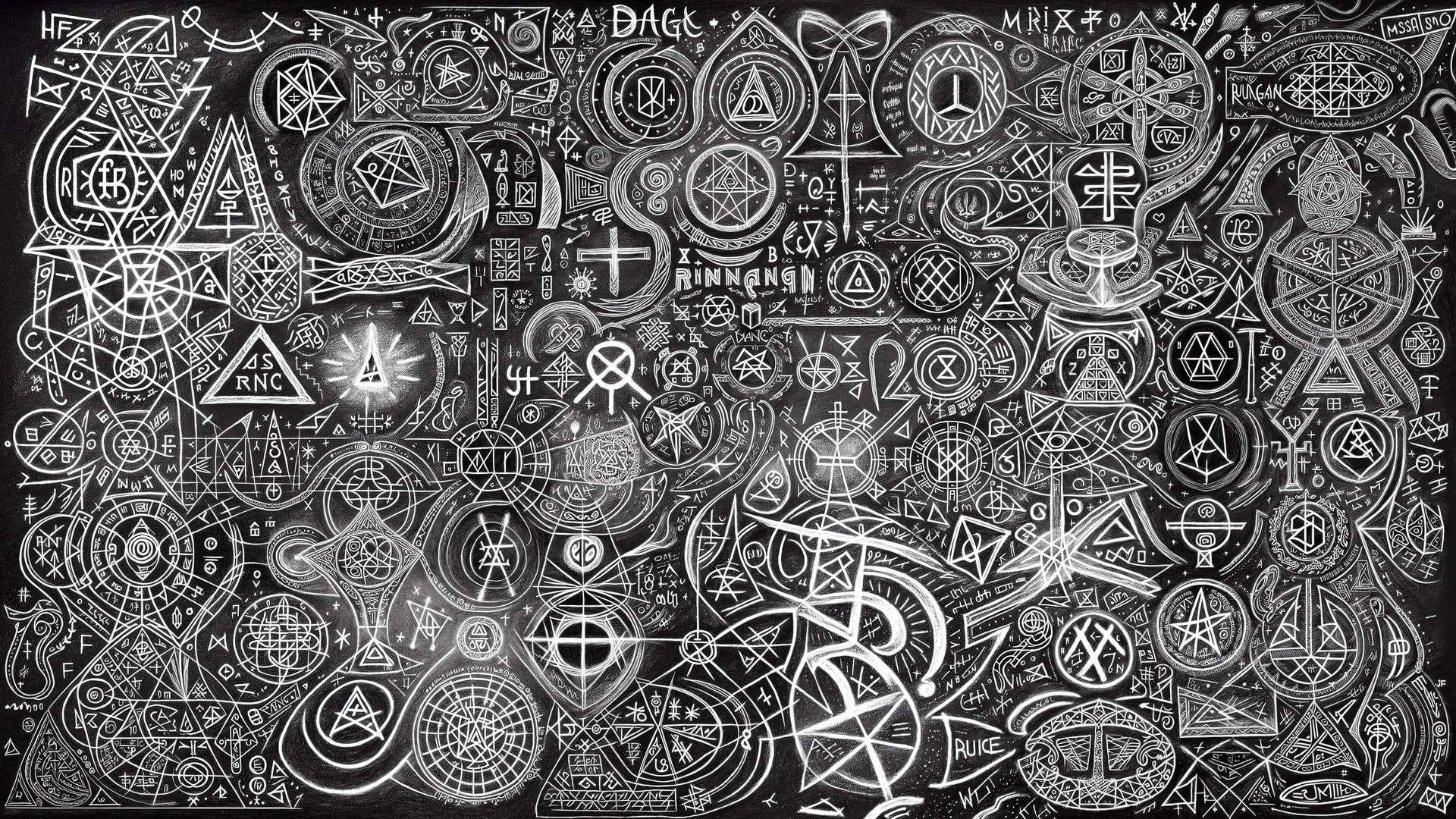}
    \caption{4K image generated with 8x11 models arranged in a grid, all of them using the prompt "Magical diagrams and runes written with chalk on a blackboard, elegant, intricate, highly detailed, smooth, sharp focus, vibrant colors, artstation, stunning masterpiece". Full resolution image available at \url{https://albarji-mixture-of-diffusers-paper.s3.eu-west-1.amazonaws.com/4Kchalkboard.png}.}
    \label{fig:4kchalkboard}
\end{figure}

\begin{figure}
    \centering
    \includegraphics[width=\textwidth]{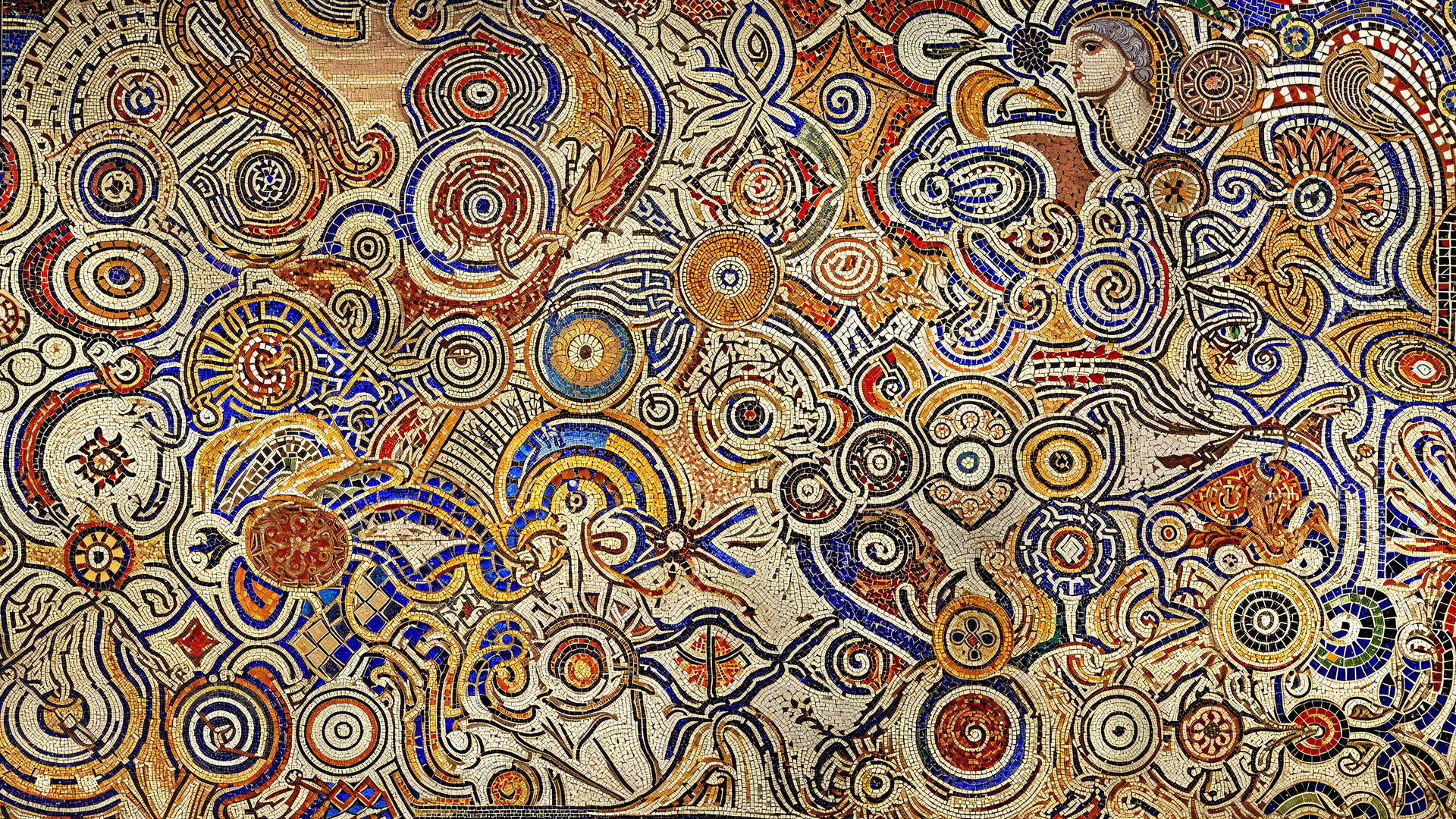}
    \caption{4K image generated with 8x11 models arranged in a grid, all of them using the prompt "Roman mosaic, elegant, intricate, highly detailed, smooth, sharp focus, vibrant colors, artstation, stunning masterpiece". Full resolution image available at \url{https://albarji-mixture-of-diffusers-paper.s3.eu-west-1.amazonaws.com/4Kmosaic.png}.}
    \label{fig:4kmosaic}
\end{figure}

Other generation samples for 4K images are presented in Figures \ref{fig:4kmiro}, \ref{fig:4kchalkboard}, \ref{fig:4kmosaic}.

\section{Additional outpainting results}
\label{app:outpainting}

\begin{figure}
    \centering
    \includegraphics[width=\textwidth]{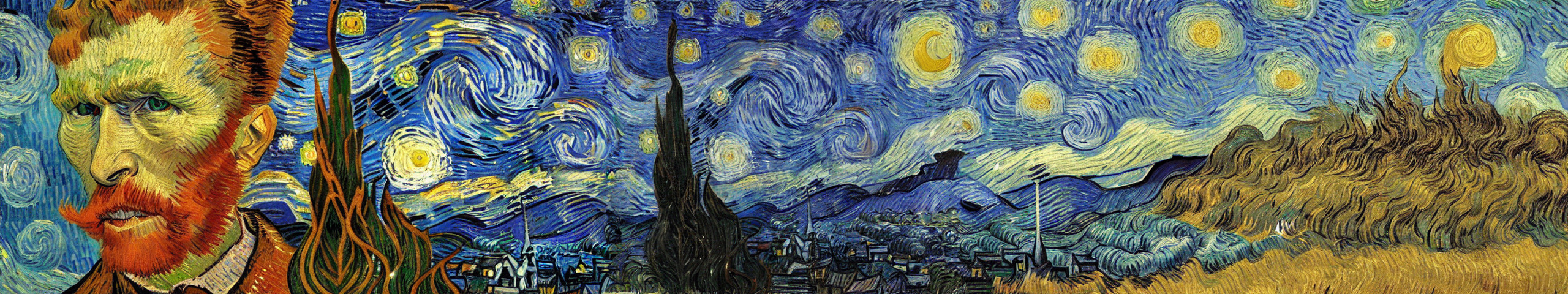}
    \caption{Mixture of Diffusers outpainting of the Starry Night by Vincent van Gogh, in both horizontal directions. The original picture is placed in the center of a $2560 \times 480$ canvas with strength $90\%$. 13 diffusion regions of size $640 \times 480$ with an overlap of $480$ columns fill the whole canvas, all of them using the prompt "by vincent van gogh, elegant, intricate, highly detailed, smooth, sharp focus, artstation, stunning masterpiece".}
    \label{fig:outpaintingStarryNight}
\end{figure}

\begin{figure}
    \centering
    \includegraphics[width=\textwidth]{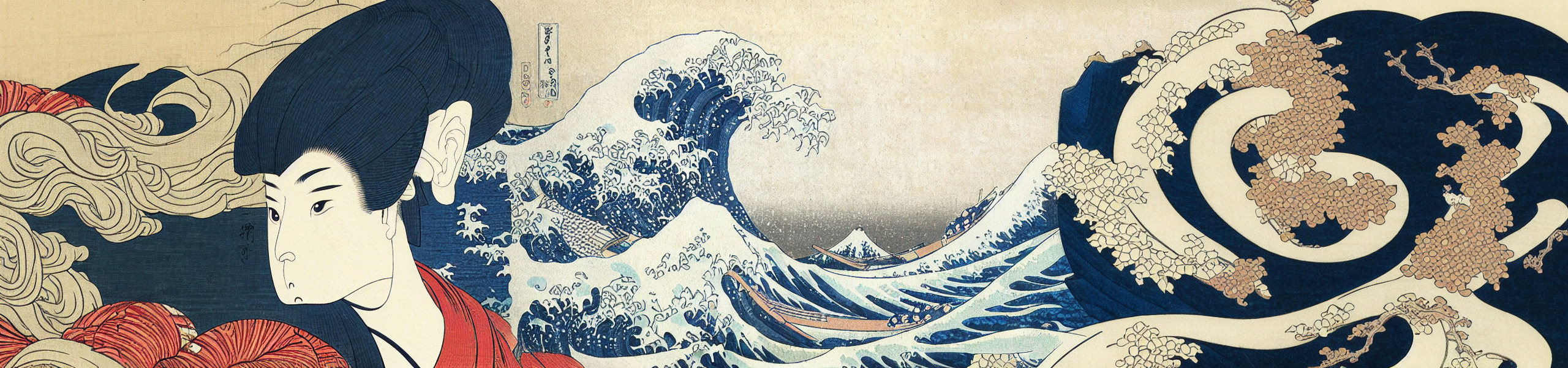}
    \caption{Mixture of Diffusers outpainting of the Great Wave off Kanagawa by Hokusai, in both horizontal directions. The original picture is placed in the center of a $2560 \times 600$ canvas with strength $75\%$. 13 diffusion regions of size $640 \times 600$ with an overlap of $480$ columns fill the whole canvas, all of them using the prompt "ukiyo-e by Hokusai, elegant, intricate, highly detailed, smooth, sharp focus, artstation, stunning masterpiece".}
    \label{fig:outpaintingWave}
\end{figure}

We have used Mixture of Diffusers to replicate the uncropping (outpainting) experiments presented in \cite{saharia2022palette}, where a guide image is extended in both horizontal directions. Figures \ref{fig:outpaintingStarryNight} and \ref{fig:outpaintingWave} show two examples of this technique.

\section{Mean of the diffusion process posterior for a noise-predicting model}
\label{app:posteriorMeanDerivation}

Although this derivation is hinted at in \cite{ho2020denoising}, we provide it here fully for the sake of clarity and completeness.

From Equation \ref{eq:xtfromeps} we have that $x_0 = (x_t - \sqrt{1 - \bar{\alpha}_t} \epsilon_{\theta}(x_t)) / \sqrt{\bar{\alpha}_t}$. Using this in Equation \ref{eq:ddpmposteriormean} we then derive

\begin{align*}
    \tilde{\mu}_t(x_t, x_0) &= \frac{\sqrt{{\bar \alpha}_{t-1}} \beta_t}{1 - {\bar \alpha}_t} x_0 + \frac{\sqrt{\alpha_t} (1 - {\bar \alpha}_{t-1})}{1 - {\bar \alpha}_t} x_t, \\
    &= \frac{\sqrt{{\bar \alpha}_{t-1}} \beta_t}{1 - {\bar \alpha}_t} ((x_t - \sqrt{1 - \bar{\alpha}_t} \epsilon_{\theta}(x_t), t) / \sqrt{\bar{\alpha}_t}) + \frac{\sqrt{\alpha_t} (1 - {\bar \alpha}_{t-1})}{1 - {\bar \alpha}_t} x_t, \\
    &= \left( \frac{\sqrt{{\bar \alpha}_{t-1}} \beta_t}{(1 - {\bar \alpha}_t) \sqrt{{\bar \alpha}_t}} + \frac{\sqrt{\alpha_t} (1 - {\bar \alpha}_{t-1}))}{1 - {\bar \alpha}_t} \right) x_t
    - \frac{\sqrt{{\bar \alpha}_{t-1}} \beta_t}{1 - {\bar \alpha}_t} \frac{\sqrt{1 - {\bar \alpha}_t}}{\sqrt{{\bar \alpha}_t}}  \epsilon_{\theta}(x_t, t), \\
    &= \frac{1}{\sqrt{\alpha_t}} \left( 
        \left( \frac{\sqrt{{\bar \alpha}_{t-1}} \beta_t}{(1 - {\bar \alpha}_t) \sqrt{{\bar \alpha}_{t-1}}} + \frac{1 - {\bar \alpha}_{t-1}}{1 - {\bar \alpha}_t} \right) x_t
        - \frac{\sqrt{{\bar \alpha}_{t-1}} \beta_t}{1 - {\bar \alpha}_t} \frac{\sqrt{1 - {\bar \alpha}_t}}{\sqrt{{\bar \alpha}_{t-1}}}  \epsilon_{\theta}(x_t, t), 
        \right) \\
    &= \frac{1}{\sqrt{\alpha_t}} \left( 
        \left( \frac{\beta_t}{1 - {\bar \alpha}_t} + \frac{1 - {\bar \alpha}_{t-1}}{1 - {\bar \alpha}_t} \right) x_t
        - \frac{\beta_t \sqrt{1 - {\bar \alpha}_t}}{1 - {\bar \alpha}_t} \epsilon_{\theta}(x_t, t)
    \right) , \\
    &= \frac{1}{\sqrt{\alpha_t}} \left(
        \frac{1 - \alpha_t + 1 - {\bar \alpha}_{t-1}}{1 - {\bar \alpha}_t}  x_t
        - \frac{(1 - \alpha_t) \sqrt{1 - {\bar \alpha}_t}}{1 - {\bar \alpha}_t} \epsilon_{\theta}(x_t, t) 
    \right) , \\
    &= \frac{1}{\sqrt{\alpha_t}} \left(
        x_t
        - \frac{1 - \alpha_t }{\sqrt{1 - {\bar \alpha}_t}} \epsilon_{\theta}(x_t, t) 
    \right) .
\end{align*}

\end{document}